\documentclass[lettersize,journal]{IEEEtran}

\usepackage{cite}
\usepackage{amsmath,amssymb,amsfonts}
\usepackage{array}
\usepackage{graphicx}
\usepackage{textcomp}
\usepackage{stfloats} 
\usepackage{url}
\usepackage{multirow}
\usepackage{booktabs}
\usepackage{balance}
\usepackage[colorlinks,urlcolor=blue,linkcolor=blue,citecolor=blue]{hyperref}

\begin{document}

\title{Spectral Adaptive Graph Transformer for Efficient and Scalable Traffic Forecasting}

\author{Amirhossein Ghaffari, \IEEEmembership{Member, IEEE}, Susanna Pirttikangas, \IEEEmembership{Senior Member, IEEE}, and Ekaterina Gilman, \IEEEmembership{Member, IEEE}

\thanks{This work was supported by the Emerging Projects program, Infotech Oulu; the CO2CREATION Strategic Research Council project (grant 372355), by the EC through the HEU NEUROCLIMA project (GA 101137711) and the HEU ARGENTIC project (GA 101298599), the Interreg Aurora ResilientEdge project (Grant Number: 20373282), the ERDF (project numbers A81568, A91867), and by the Finnish Innovation Fund Sitra through the Physical-AI project (agreement 640102).}

\thanks{Amirhossein Ghaffari, Susanna Pirttikangas and Ekaterina Gilman are with the Future Computing Group, University of Oulu, 90570 Oulu, Finland (e-mail: amirhossein.ghaffari@oulu.fi, 
susanna.pirttikangas@oulu.fi,  ekaterina.gilman@oulu.fi).}

}

\markboth{}
{Ghaffari \MakeLowercase{\textit{et al.}}: Spectral Adaptive Graph Transformer for Efficient and Scalable Traffic Forecasting}

\maketitle
\begin{abstract}
Traffic forecasting models based on spectral graph convolution offer an interpretable account of how information is mixed between sensors, but they rely on an eigendecomposition of the graph Laplacian. It costs cubic time, is fixed by the road network, and provides a symmetric operator that cannot represent the directed propagation of traffic congestion. Attention-based spatio-temporal transformers remove these limitations but generate a dense sensor-by-sensor attention map at each layer and time step, so memory, rather than accuracy, becomes the binding constraint on large networks. We introduce the spectral adaptive graph transformer, whose spatial operator is learned rather than derived from the Laplacian, directed rather than symmetric, and conditioned on the current traffic state rather than a static state, at a cost linear in the number of sensors. A sparse road graph diffusion branch provides the strictly local propagation that a low-rank operator cannot express, and a zero-initialized gate learns the balance between the two. Before spatial mixing, a unit-mean-gain frequency-bin attention module reweights each sensor's short-term periodicities according to its identity and current context. Across six public benchmarks spanning 170 to 1843 sensors, evaluated over three random seeds, the proposed model attains the lowest error of any single end-to-end model on one benchmark and remains competitive with the strongest current attention models on the others, while using about a quarter of their memory and training about twice as fast.
\end{abstract}


\section{Introduction}
\label{sec:intro}

\IEEEPARstart{T}{raffic} forecasting underpins congestion
management, adaptive signal control, and route guidance. The problem is difficult because the signal is jointly structured in time and space: a reading evolves according to its own history, the conditions propagating along the road network, and strong daily and weekly periodicity. Over the past decade, the field has converged on two broad architectural strategies, each of which trades one difficulty for another.

The first approach models spatial dependence with \emph{graph convolution}. Building on spectral graph theory~\cite{shuman2013emerging,defferrard2016convolutional,kipf2016semi}, methods such as DCRNN~\cite{li2017diffusion}, STGCN~\cite{yu2017spatio} and Graph WaveNet~\cite{wu2019graph} propagate information along a predefined or learned adjacency. StemGNN~\cite{cao2020spectral} takes the spectral view furthest, performing inter-series mixing entirely in the graph-Fourier domain. This is appealing; filtering in the spectral domain has a clean interpretation as selectively attenuating spatial modes, but it inherits three constraints of the classical construction. The graph-Fourier basis is the eigenbasis of the Laplacian, so obtaining it costs $\mathcal{O}(N^3)$ for $N$ sensors; the basis is fixed by the graph and cannot adapt to the data; and because an orthogonal basis is applied symmetrically, the resulting operator is symmetric and cannot represent directed influence, even though congestion demonstrably propagates upstream-to-downstream along one-way roads.

The second strategy replaces the graph operator with \emph{attention}. Following the Transformer~\cite{vaswani2017attention}, models such as GMAN~\cite{zheng2020gman}, PDFormer~\cite{jiang2023pdformer}, STAEformer~\cite{liu2023staeformer}, and TESTAM~\cite{lee2024testam} enable every sensor to attend to every other sensor, learning the interaction structure directly from data. STAEformer showed that a vanilla Transformer, paired with a learned spatio-temporal adaptive embedding, is sufficient to achieve state-of-the-art accuracy, indicating that much of the benefit comes from flexible representation rather than elaborate graph machinery ~\cite{liu2023staeformer}. The cost, however, is quadratic: an $N\times N$ attention map is materialized at every layer and time step. On the benchmarks that dominate the literature, typically $200$ to $900$ sensors, this is tolerable. On networks approaching a few thousand sensors, it becomes the binding constraint, and accuracy ceases to be the quantity that determines whether a model can be used at all.

This tension motivates our work. We investigate whether the interpretability and structure of spectral mixing can be retained while removing both the eigendecomposition of the spectral approach and the quadratic cost of the attention approach. We propose the Spectral Adaptive Graph Transformer (SAGT) model as a solution. Comparing the proposed model with the classical spectral form makes three clear distinctions: 1) the fixed eigenbasis becomes a learned, truncated basis, thereby removing the eigendecomposition; 2) the static filter becomes input-conditioned; and 3) the symmetric sandwich becomes a directed pair, allowing upstream and downstream influences to differ. As the operator is never materialized, the computation passes through a lower-dimensional bottleneck; therefore, its cost is linear rather than quadratic in the number of nodes.

Two further observations shaped the design of SAGT. First, low-rank global mixing cannot, by construction, express strictly local propagation along an individual road segment. Therefore, we pair it with an inexpensive sparse diffusion branch on the road graph and let a zero-initialized gate learn how much of each to use. Second, sensors differ in how informative their \emph{within-window} variation is: e.g., on a busy corridor, the movement over the last hour is largely signal, whereas on a lightly traveled ramp, much of it is measurement noise. We therefore introduce a per-sensor attention mechanism over the discrete Fourier bins of the input window that amplifies or suppresses individual oscillation rates before any spatial mixing occurs. Longer rhythms, such as the daily and weekly cycles, are outside this window and are carried by the calendar embeddings instead. Unlike frequency-domain forecasters such as FEDformer~\cite{zhou2022fedformer} and Autoformer~\cite{wu2021autoformer}, which operate on the series as a whole, the proposed mechanism is conditioned jointly on sensor identity and current context, and its attention weights are directly inspectable.

Our contributions are as follows:

\begin{itemize}
\item We introduce a spatial operator that is simultaneously \emph{learned} (requiring no eigendecomposition), \emph{directed} (asymmetric, with sign-unconstrained filter coefficients), and \emph{input-conditioned} (recomputed per time step), at a cost linear in the number of nodes rather than quadratic.

\item We propose unit-mean gain frequency-bin attention, a per-sensor reweighting of the input window's discrete Fourier bins. We argue that applying the softmax weights directly is a poor parameterization rather than a poor design: at initialization, it attenuates the reconstructed measurement channel by a factor equal to the number of bins relative to the other components of the token, so the mechanism must first undo its own attenuation before it can help. Rescaling so that uniform attention is the exact identity removes the problem.

\item We evaluate eight baselines and one composite method across six datasets, spanning $170$ to $1843$ sensors, using a single shared data pipeline, and report every entry as the mean and standard deviation over three seeds. Relative to the dense-attention models that define the accuracy frontier on these benchmarks, SAGT is the most accurate on PEMS08 and within $0.13$~MAE of the leader on three of the remaining five benchmarks, while using roughly a quarter of the memory at equal batch size, and remains trainable at scales where they exhaust GPU memory.

\end{itemize}

The remainder of this paper is organized as follows. Section~\ref{sec:related} reviews the related work and situates the work, Section~\ref{sec:method} presents the model, Section~\ref{sec:results} reports the experiments and provides discussions, and Section~\ref{sec:conclusion} concludes the work.

\section{Related Work}
\label{sec:related}

\subsection{Graph-Based Spatio-Temporal Forecasting}
\label{sec:rel-graph}

Modern traffic forecasting considers sensors not as independent series but as nodes of a road network. DCRNN~\cite{li2017diffusion} models spatial dependence as a diffusion process on a directed graph, coupled with a sequence-to-sequence recurrent decoder; STGCN~\cite{yu2017spatio} instead interleaves Chebyshev graph convolutions~\cite{defferrard2016convolutional} with temporal convolutions, avoiding recurrence entirely; and T-GCN~\cite{zhao2019tgcn} takes the most direct composition of all, applying a two-layer graph convolution to obtain spatial features and a gated recurrent unit to propagate them in time. All three models construct their adjacency from road distances. The T-GCN authors report a specific and reproducible failure: the model predicts poorly at local minima and maxima, which they attribute to the graph convolution acting as a smooth filter in the Fourier domain, flattening peaks as it is applied \cite{zhao2019tgcn}. We read this as a statement about the operator rather than about the temporal backbone. A filter that is symmetric and non-negative can attenuate spatial modes, but cannot form contrasts between them, and the sharp extremes of a traffic profile are precisely the structure such an operator removes. 

Subsequent work relaxed the dependence on a predefined graph. Graph WaveNet~\cite{wu2019graph} augments the physical adjacency with a self-adaptive matrix formed from learned node embeddings, and AGCRN~\cite{bai2020agcrn} dispenses with the predefined graph altogether, generating node-specific convolution parameters from embeddings. ASTGCN~\cite{guo2019attention} and STSGCN~\cite{song2020stsgcn} add attention and synchronous spatio-temporal blocks, respectively, while DGCRN~\cite{li2023dgcrn} generates a dynamic graph at each step from the current hidden state, recognizing that interaction structure varies with traffic conditions.

DSTAGNN~\cite{lan2022dstagnn} pushes this line furthest on the side of graph construction: it derives a spatio-temporal aware graph from Wasserstein distances between the daily traffic distributions of node pairs, so that two sensors are related when their behavior is similar rather than when they are physically close, and it modulates each order of a Chebyshev expansion with a separate attention head. This idea reinforces the premise underlying our own design, that useful spatial structure is better discovered from data than inherited just from road distances. 

Another concern is cost. MTGODE~\cite{jin2022multivariate} reformulates graph propagation and temporal aggregation as two coupled neural ODEs, learns the graph structure jointly rather than assuming it, and reports lower parameter counts and memory than the discrete architectures it generalizes, because propagation depth is no longer tied to a stack of individually parameterized layers. Its objective coincides with ours, accuracy that need not be purchased with parameters or stored activations, but the resolution differs in direction. MTGODE refines the spatial operator along \emph{depth}, taking the integration step to zero; SAGT constrains it along \emph{width}, forcing all inter-sensor communication through a rank-$k$ bottleneck so that no $N\times N$ quantity is ever formed.

SAGT is motivated by the following three observations. Learned interaction structure outperforms purely distance-based adjacency, motivating learning the spectral basis; dynamic, condition-dependent structure outperforms static structure, motivating conditioning the spectral filter on the current input; and the cost of obtaining a model is a design constraint rather than an afterthought, motivating the efficiency analysis. We retain the physical road graph, but only as a sparse local branch rather than as the principal mixing mechanism.

\subsection{Spectral and Frequency-Domain Approaches}
\label{sec:rel-spectral}

Spectral graph theory provides the formal foundation for graph convolution~\cite{defferrard2016convolutional}. Early spectral networks filtered directly in the Laplacian eigenbasis; ChebNet~\cite{defferrard2016convolutional} replaced the explicit eigendecomposition with a Chebyshev polynomial expansion, and GCN~\cite{kipf2016semi} simplified this to a first-order approximation. These two approximations remain the workhorses of the field: they are what STGCN, T-GCN, and DSTAGNN ultimately apply to, regardless of the graph they are given. The tension these works negotiate, between the expressiveness of an explicit spectral basis and the $\mathcal{O}(N^3)$ cost of obtaining it, is precisely the tension SAGT addresses, but with a different resolution: rather than approximating the analytic basis with polynomials, we learn a low-rank orthonormal basis directly.

A related restriction appears from the opposite direction in analyses of deep propagation. Repeatedly applying a fixed normalized diffusion operator drives the graph spectrum towards a degenerate limit~\cite{jin2022multivariate}, which is the spectral statement of over-smoothing: a non-negative filter can only remove energy from spatial modes, never redistribute it. Our filter coefficients are deliberately left unsquashed and may therefore be negative (Section~\ref{sec:spatial}), so the operator is indefinite and can express both contrast and smoothing.

StemGNN~\cite{cao2020spectral} is the closest prior work. It performs inter-series mixing via a graph Fourier transform and intra-series modeling via a discrete Fourier transform, operating entirely in the spectral domain. SAGT departs from it in four respects: the basis is learned rather than derived from a latent-correlation Laplacian; it is truncated to rank $k\ll N$, making cost linear in $N$; the analysis and synthesis bases differ, making the operator directed; and the filter is conditioned on the current input rather than fixed after training.

In the broader time-series literature, frequency-domain methods have been used to control the cost of long-sequence modeling. Autoformer~\cite{wu2021autoformer} introduces an auto-correlation mechanism with series decomposition, and FEDformer~\cite{zhou2022fedformer} performs attention on a randomly selected subset of Fourier modes. These operate on the temporal frequency content of a series as a whole. Our frequency-bin attention differs in granularity and purpose: it is applied per sensor, conditioned jointly on a learned sensor identity and the current window, and its role is to reweight periodicities before spatial mixing rather than to reduce sequence-length complexity.

\subsection{Attention-Based Traffic Models}
\label{sec:rel-attention}

Following the Transformer~\cite{vaswani2017attention}, attention has progressively displaced explicit graph convolution. GMAN~\cite{zheng2020gman} applies spatial and temporal attention in an encoder--decoder with a transform layer between them. PDFormer~\cite{jiang2023pdformer} models propagation delay explicitly, using geographic and semantic masked attention with $k$-Shape clustering of traffic patterns. TESTAM~\cite{lee2024testam} adopts a mixture-of-experts formulation in which identity, adaptive-graph, and attention experts are combined by a memory-based gate trained with auxiliary routing losses. MegaCRN~\cite{jiang2023spatio} learns a meta-graph via a memory bank, with a disentangling objective over prototypes.

The question of how graph structure should be incorporated into an attention model has been examined most directly outside the context of traffic forecasting. Graphormer~\cite{ying2021transformers} shows that a standard Transformer becomes competitive with, and on several benchmarks superior to, dedicated graph networks once structural information is injected as bias, degree-based centrality added to node features, and shortest-path distance added to the attention logits, rather than as a convolution operator. Two observations carry over to our setting. First, structure need not be imposed through a propagation rule at all; it can live in the representation, which is what STAEformer's adaptive embedding does implicitly and what our learned spatial modes do explicitly. Second, the authors identify the quadratic cost of self-attention as the property that limits the architecture's scalability on large graphs and identify efficient variants as the required next step. Traffic networks are exactly the regime in which that restriction becomes binding, since the spatial axis is the long one.

STAEformer~\cite{liu2023staeformer} is of particular relevance. Its central finding is that a vanilla Transformer, combined with a learned spatio-temporal adaptive embedding, matches or exceeds the performance of far more elaborate architectures. We adopt this embedding and the temporal attention backbone in our design. Our main contribution is orthogonal: we replace the spatial half of that architecture, which is the component responsible for its quadratic cost.

\subsection{Mixture-of-Experts and Composite Forecasters}
\label{sec:rel-moe}

A distinct line of work improves accuracy not by designing a better single predictor but by \emph{routing} among several predictors. The two differ in an important respect. In TESTAM~\cite{lee2024testam}, the experts are internal modules trained jointly with the gate, so the result is still one model with one training run. In GC-MoE~\cite{gcmoe}, the experts are complete, independently pretrained forecasters that are loaded frozen and held in evaluation mode. Only a graph-conditioned router is trained: it reads the input window together with a neighborhood-aggregated context from the road graph and emits a mixture weight over experts per sensor, which is combined with a bounded output-refinement correction and regularised by load-balancing and routing-entropy terms. The router is lightweight, but the deployable system comprises the router and every expert.

We include GC-MoE in our comparison because it is a strong point of reference on these benchmarks, but we report it separately and do not rank it against the single end-to-end models. Two reasons make a direct ranking misleading. Its capacity is the union of several forecasters rather than that of one architecture, and its true training cost is dominated by the experts rather than by the router. This distinction is orthogonal to our contribution rather than in competition with it: routing operates over predictors, whereas SAGT is a predictor, and a cheaper single model of the kind we propose is a better expert and a better standalone forecaster.

\subsection{Efficient Attention}
\label{sec:rel-efficient}

The quadratic cost of self-attention has been attacked extensively in the general sequence-modeling literature. Linformer~\cite{wang2020linformer} projects keys and values to a lower-dimensional space; Performer~\cite{choromanski2020rethinking} approximates the softmax kernel with random features; Informer~\cite{zhou2021informer} sparsifies attention by selecting dominant queries. These methods reduce complexity along the sequence axis, which in traffic forecasting is short ($T=12$) and therefore not the bottleneck.

In traffic forecasting, STWave~\cite{fang2023spatio} attacks the same spatial axis, disentangling trends from events and replacing full graph attention with a query-sampling scheme that reduces cost to $\mathcal{O}(N\log N)$ while retaining a global receptive field. Its graph-wavelet positional encoding still requires an $\mathcal{O}(N^{3})$ Laplacian eigendecomposition as a preprocessing step, which is precisely the dependency our learned basis removes.

SAGT addresses the spatial axis, whose length equals the number of sensors. This is the same axis that graph transformers identify as their own scaling barrier~\cite{ying2021transformers}. Rather than approximating a softmax attention map, we replace it with an explicitly low-rank, learned spectral operator. This gives an exact rather than approximate computation, retains a spectral interpretation of what the operator does, and, because the operator is never materialized, avoids storing any $N\times N$ activation.

\subsection{Benchmarking and Reproducibility}
\label{sec:rel-bench}

Comparisons in this field are complicated by inconsistent preprocessing, splits and metric definitions. Standardized suites such as LibCity~\cite{wang2021libcity} and BasicTS~\cite{shao2024exploring} have documented substantial deviations from originally published numbers when models are re-evaluated under a unified protocol. A further source of variation lies upstream of the model entirely: STM-Graph~\cite{stmgraph} holds the architecture fixed and varies only how raw spatio-temporal data are mapped onto a graph, and finds that the choice of partitioning and resolution moves the reported metrics by margins far larger than those separating competing models, with the ranking of metrics themselves sensitive to the resulting graph. Two consequences bear on the present work. First, an accuracy comparison is only meaningful when the graph and the pipeline that produced it are held fixed, which is why every model here consumes identical tensors from one shared pipeline. Second, the road graph is a modeling decision rather than ground truth, which is part of why SAGT treats it as one of two sparse branches rather than the sole channel for spatial information.

Our own experience, reported in Section~\ref{sec:baselines}, is consistent with the standardized suites. We therefore evaluate every model using a single shared data pipeline and a single metric implementation.

\section{Methodology}
\label{sec:method}

\subsection{Problem Formulation and Notation}
\label{sec:problem}

We consider a road network instrumented with $N$ sensors, whose topology is encoded by an adjacency matrix $\mathbf{A}\in\mathbb{R}^{N\times N}$, where $A_{ij}>0$ indicates a directed road connection from sensor $i$ to sensor $j$. At each discrete time step, every sensor reports a traffic measurement (flow or speed, depending on the dataset) together with two calendar variables.

Let $T$ denote the observed input window length and $H$ the forecast horizon. For a batch of $B$ samples, the input tensor is

\begin{equation}
\mathcal{X}\in\mathbb{R}^{B\times T\times N\times C},\qquad C=3,
\label{eq:input}
\end{equation}

whose channels are the traffic measurement $x^{\mathrm{val}}$, the time-of-day fraction $x^{\mathrm{tod}}\in[0,1)$, and the day-of-week index $x^{\mathrm{dow}}\in\{0,\dots,6\}$. The task is to predict $\hat{\mathcal{Y}}\in\mathbb{R}^{B\times H\times N\times 1}$ from $\mathcal{X}$. We use $b$ for the batch index, $t$ for time, $v$ for the sensor (node), and $j$ for a spectral or frequency index. Table~\ref{tab:notation} lists all symbols with the values used in our experiments.

The measurement channel is standardized using statistics computed on the training split only, so that no information leaks from validation or test data:

\begin{equation}
\tilde{x}^{\mathrm{val}}=\frac{x^{\mathrm{val}}-\mu}{\sigma},
\label{eq:zscore}
\end{equation}

where $\mu$ and $\sigma$ are the mean and standard deviation of the training measurements. The model operates in this normalized space, and predictions are mapped back by $\hat{y}=\sigma\hat{y}^{\mathrm{norm}}+\mu$ before any loss or metric is computed, so all reported errors are in physical units.

\begin{table}[!t]
\renewcommand{\arraystretch}{1.15}
\caption{Notation and Configuration Used}
\label{tab:notation}
\centering
\footnotesize
\begin{tabular}{llc}
\toprule
Symbol & Meaning & Value / shape \\
\midrule
\multicolumn{3}{l}{\emph{Sizes and hyperparameters}}\\
$N$    & number of sensors (nodes)                & 170--1843 \\
$T$    & input window length                      & 12 \\
$H$    & forecast horizon                         & 12 \\
$C$    & input channels                           & 3 \\
$d$    & model (token) width                      & 152 \\
$d_f$  & feed-forward hidden width                & 256 \\
$L$    & number of stacked blocks                 & 3 \\
$h$    & attention heads                          & 4 \\
$d_h$  & per-head width, $d/h$                    & 38 \\
$k$    & spectral rank (retained modes)           & 32 \\
$F$    & DFT bins, $\lfloor T/2\rfloor+1$         & 7 \\
$P$    & time steps per day                       & 288 / 144 \\
$d_e$  & node-embedding width, freq.\ attention   & 16 \\
$p$    & dropout probability                      & 0.1 \\
$|E|$  & road edges (non-zeros of $\mathbf{A}$)   & $\mathcal{O}(10^3)$ \\
$\ell$ & block index, $1\le\ell\le L$             & --- \\
\midrule
\multicolumn{3}{l}{\emph{Data and token tensors}}\\
$\mathcal{X}$ & input tensor, \eqref{eq:input}          & $B{\times}T{\times}N{\times}C$ \\
$\mathbf{s}_{b,v}$ & standardized window of sensor $v$, \eqref{eq:sdef} & $T$ \\
$\mathcal{Z}^{(\ell)}$ & token tensor after block $\ell$, \eqref{eq:ztensor} & $B{\times}T{\times}N{\times}d$ \\
$\mathbf{Z}_{b,:,v}$ & temporal slice (one sensor)       & $T{\times}d$ \\
$\mathbf{Z}_{b,t}\!\equiv\!\mathbf{Z}_{b,t,:}$ & spatial slice (one step) & $N{\times}d$ \\
$\mathbf{z}_{b,t,v}$ & one token, row $v$ of $\mathbf{Z}_{b,t}$ & $d$ \\
\midrule
\multicolumn{3}{l}{\emph{Spatial layer}}\\
$\mathbf{U}_q,\mathbf{U}_k$ & learned orthonormal bases, \eqref{eq:qr} & $N{\times}k$ \\
$\boldsymbol{\theta}_{b,t}$ & input-conditioned filter, \eqref{eq:theta} & $k$ \\
$\mathbf{M}_{b,t}$ & effective operator, rank $\le k$, \eqref{eq:operator} & $N{\times}N$ \\
$\mathbf{G}^{\mathrm{glo}}_{b,t}$ & global branch output, \eqref{eq:synthesis} & $N{\times}d$ \\
$\mathbf{G}^{\mathrm{loc}}_{b,t}$ & local branch output, \eqref{eq:local} & $N{\times}d$ \\
$\mathbf{G}^{\mathrm{fus}}_{b,t}$ & fused branch output, \eqref{eq:fuse} & $N{\times}d$ \\
\bottomrule
\end{tabular}
\end{table}

\subsection{Design Rationale}
\label{sec:rationale}

Two model families dominate current traffic forecasting models' components, and each carries a structural limitation that motivates our design.

\emph{Spectral graph models}, exemplified by StemGNN \cite{cao2020spectral}, mix information across sensors in the graph-Fourier domain. Given the symmetrically normalized Laplacian $\mathbf{L}_{\mathrm{sym}}=\mathbf{I}-\mathbf{D}^{-1/2}\mathbf{A}\mathbf{D}^{-1/2}$ with eigendecomposition $\boldsymbol{\Phi}\boldsymbol{\Lambda}\boldsymbol{\Phi}^{\!\top}$, a spectral filter acts as

\begin{equation}
g_{\boldsymbol{\theta}}(\mathbf{L}_{\mathrm{sym}})\mathbf{x} =\boldsymbol{\Phi}\,g_{\boldsymbol{\theta}}(\boldsymbol{\Lambda})\,\boldsymbol{\Phi}^{\!\top}\mathbf{x},
\label{eq:classical}
\end{equation}

where $\mathbf{D}=\operatorname{diag}(\mathbf{A}\mathbf{1})$ is the degree matrix, $\mathbf{x}\in\mathbb{R}^{N}$ is a graph signal holding one value per sensor, the columns of $\boldsymbol{\Phi}\in\mathbb{R}^{N\times N}$ are the graph-Fourier modes, $\boldsymbol{\Lambda}=\operatorname{diag}(\lambda_1,\dots,\lambda_N)$ contains the corresponding graph frequencies, and $g_{\boldsymbol{\theta}}(\cdot)$ is a filter with learnable parameters $\boldsymbol{\theta}$ applied to each eigenvalue~\cite{shuman2013emerging,defferrard2016convolutional}. The product $\boldsymbol{\Phi}^{\!\top}\mathbf{x}$ is the graph Fourier transform of $\mathbf{x}$, the filter rescales each spectral component, and multiplication by $\boldsymbol{\Phi}$ maps the result back to the sensors.

This formulation is elegant but has three practical drawbacks. Obtaining $\boldsymbol{\Phi}$ requires an eigendecomposition costing $\mathcal{O}(N^3)$. The basis is \emph{fixed}, so it cannot adapt to data. Since $\boldsymbol{\Phi}$ is orthogonal and the filter is applied symmetrically, the resulting operator is symmetric and therefore \emph{cannot express directed influence}, yet traffic propagates asymmetrically, from upstream to downstream.

\emph{Attention-based spatio-temporal transformers}, exemplified by STAEformer \cite{liu2023staeformer}, replace the graph operator with dense self-attention over sensors. This is expressive and adaptive, but materializes an $N\times N$ attention map in every layer at every time step, giving $\mathcal{O}(N^2)$ memory and compute. On large networks, this becomes the binding constraint rather than accuracy.

SAGT retains the desirable properties of both while removing these obstacles. Its spatial operator is (i) \emph{learned} rather than obtained by eigendecomposition, (ii) \emph{directed} and sign-unconstrained rather than symmetric positive semi-definite, (iii) \emph{conditioned on the current input} rather than static, and (iv) of cost $\mathcal{O}(Nk+|E|)$ rather than $\mathcal{O}(N^2)$, with $k\ll N$. The temporal pathway reuses the adaptive embedding and vanilla transformer of STAEformer, which we found to be a strong and inexpensive backbone.

\begin{figure*}[!t]
\centering
\includegraphics[width=1\textwidth]{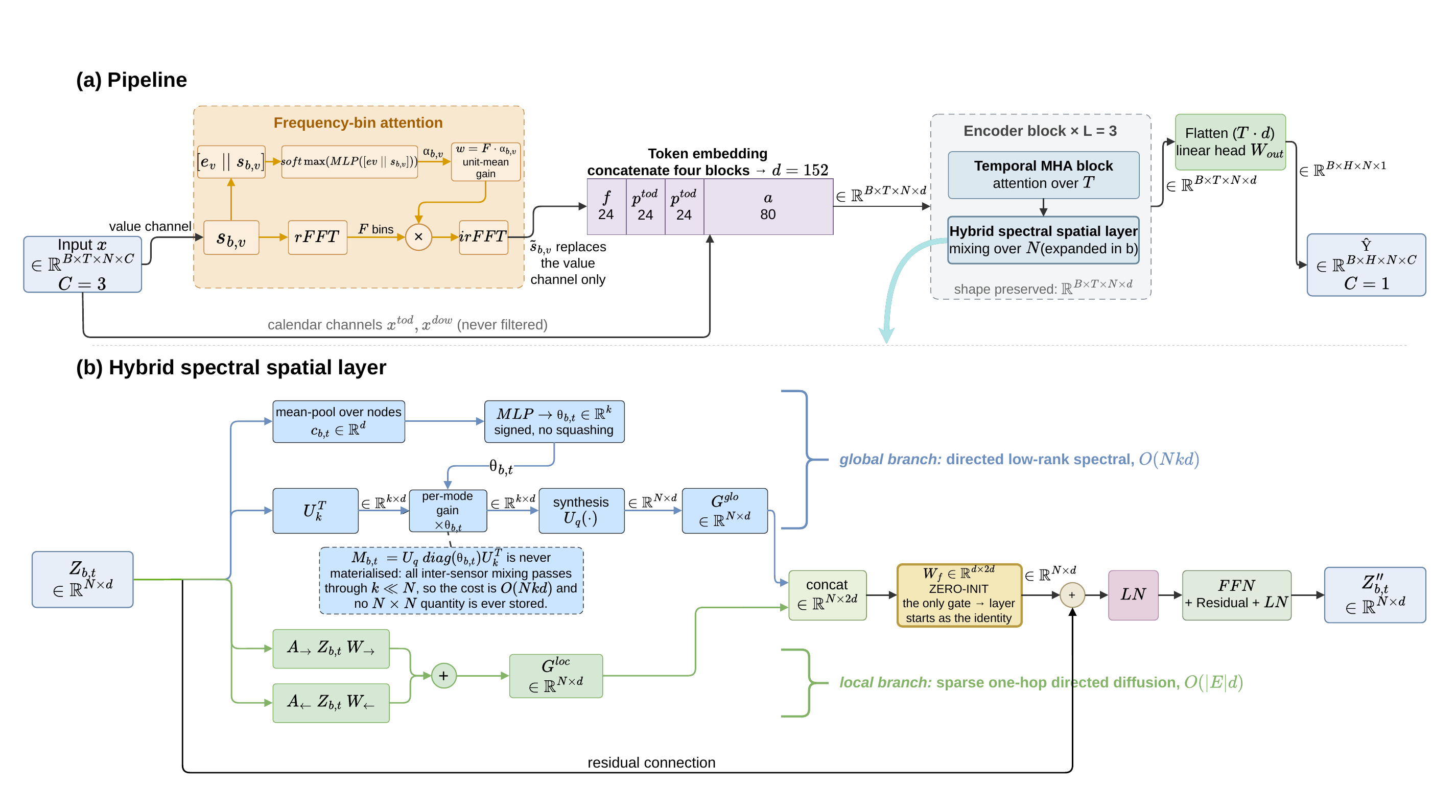}
\caption{Architecture of SAGT. (a) Pipeline. The input $\mathcal{X}$ passes through frequency-bin attention (Sec.~\ref{sec:freqattn}), whose sole output is the reweighted series that substitutes for the value channel; the calendar channels are never filtered. Tokens are formed by concatenating four blocks of width $24/24/24/80$ into $d=152$ (Sec.~\ref{sec:embedding}), refined by $L=3$ blocks of a temporal attention layer (Sec.~\ref{sec:temporal}) followed by a hybrid spatial layer (Sec.~\ref{sec:spatial}), and mapped to all $H$ horizons by one linear head. Shape is preserved throughout the encoder. (b) The hybrid spatial layer, expanded. The global branch analyses with $\mathbf{U}_k^{\top}$, applies the input-conditioned signed filter $\boldsymbol{\theta}_{b,t}$ predicted from the sensor-pooled context, and synthesizes with $\mathbf{U}_q\!\neq\!\mathbf{U}_k$; the local branch applies two sparse directed diffusions on the road graph. The effective operator $\mathbf{M}_{b,t}$ is never formed: all inter-sensor mixing passes through the rank-$k$ waist, so no $N\times N$ quantity is stored. The fusion projection $\mathbf{W}_f$ is the single zero-initialized gate, so the layer begins as the identity.}
\label{fig:arch}
\end{figure*}

Fig.~\ref{fig:arch} summarizes the resulting architecture and fixes the notation used throughout this section. Panel~(a) traces one input window from $\mathcal{X}$ to $\hat{\mathcal{Y}}$ and makes explicit that the frequency-bin attention of Section~\ref{sec:freqattn} enters the network at exactly one point, as a replacement for the value channel in the token construction of Section~\ref{sec:embedding}, and nowhere else. Panel~(b) expands the hybrid spatial layer. Read against the classical form~\eqref{eq:classical}, it makes the three substitutions visible: the fixed eigenbasis is replaced by the learned pair $(\mathbf{U}_k,\mathbf{U}_q)$, the static filter by the context-conditioned $\boldsymbol{\theta}_{b,t}$, and the symmetric sandwich by a directed pair. The annotated rank-$k$ waist is where the linear cost of Section~\ref{sec:complexity} originates, and the memory measurements of Section~\ref{sec:efficiency} are a direct consequence of the fact that nothing of size $N\times N$ appears.

The following subsections follow the order in which data flows through the model.

\subsection{Frequency-Bin Attention}
\label{sec:freqattn}

Different sensors express different short-term dynamics within the observed window: some vary smoothly, while others fluctuate from one reading to the next. Rather than leaving the network to discover this implicitly, SAGT first re-weights each sensor's signal directly in the time-frequency domain. Note that the transform spans the input window only, so the bins describe within-window oscillation rates rather than daily or weekly periodicity, which the calendar embeddings of Section~\ref{sec:embedding} supply.

For sample $b$ and sensor $v$, let $\mathbf{s}_{b,v}\in\mathbb{R}^{T}$ collect that sensor's standardized measurements across the input window. Therefore, it is the sequence obtained by applying~\eqref{eq:zscore} at each of the $T$ steps as follows:

\begin{equation}
\mathbf{s}_{b,v}=\big[\tilde{x}^{\mathrm{val}}_{b,0,v},\;\dots,\; \tilde{x}^{\mathrm{val}}_{b,T-1,v}\big]^{\!\top}, \quad\text{so}\quad s_{b,v}[t]=\tilde{x}^{\mathrm{val}}_{b,t,v}.
\label{eq:sdef}
\end{equation}

Now, $\mathbf{s}_{b,v}$ gathers the $T$ readings of one sensor into the vector on which a Fourier transform can be taken. Its real discrete Fourier transform is

\begin{equation}
S_{b,v}[j]=\sum_{t=0}^{T-1}s_{b,v}[t]\,e^{-\mathrm{i}2\pi jt/T},
\quad j=0,\dots,F-1,
\label{eq:rfft}
\end{equation}

with $F=\lfloor T/2\rfloor+1$ bins. Bin $j$ corresponds to a period of $T/j$ time steps; for five-minute data with $T=12$ the bins span the mean level ($j{=}0$) down to a ten-minute oscillation ($j{=}6$), and on the EXPY-TKY dataset, whose sampling interval is ten minutes, the same highest bin corresponds to a twenty-minute oscillation.

Each sensor owns a learnable embedding $\mathbf{e}_v\in\mathbb{R}^{d_e}$, encoding persistent sensor-specific behavior. Concatenating it with the current window yields a descriptor that is both static (which sensor this is) and dynamic (what it is doing now). A two-layer perceptron maps this descriptor to a distribution over frequency bins:

\begin{equation}
\boldsymbol{\alpha}_{b,v}=\operatorname{softmax}\!\big(\mathbf{W}_2\,\rho(\mathbf{W}_1\mathbf{q}_{b,v}+\mathbf{b}_1)+\mathbf{b}_2\big),
\label{eq:freqattn}
\end{equation}

where $\mathbf{q}_{b,v}=[\mathbf{e}_v\,\|\,\mathbf{s}_{b,v}]$, $\|$ denotes concatenation, $\rho(\cdot)=\max(0,\cdot)$, $\mathbf{W}_1\in\mathbb{R}^{32\times(d_e+T)}$, $\mathbf{W}_2\in\mathbb{R}^{F\times32}$, and $\boldsymbol{\alpha}_{b,v}$ lies on the probability simplex $\Delta^{F-1}$.

A subtle but consequential point arises here. Applying $\boldsymbol{\alpha}$ directly as a multiplicative mask is a poor parameterization. Since $\sum_j\alpha_{b,v}[j]=1$, the average bin weight is $1/F$, so the reconstructed series is attenuated by roughly a factor of $F$. This is not a loss of expressiveness: the filtered channel enters the network only through the linear projection $\mathbf{W}_{\mathrm{in}}$ of Section~\ref{sec:embedding}, whose corresponding column could in principle absorb the constant. It is, however, a poor starting point for optimization. At initialization $\boldsymbol{\alpha}\approx\mathbf{1}/F$, so the measurement channel reaches the token roughly $F$ times smaller than the calendar and adaptive components it is concatenated with, and the network must first learn to undo an attenuation that the module imposed on itself before that module can contribute anything. We therefore rescale the attention into a \emph{unit-mean gain},

\begin{equation}
\mathbf{w}_{b,v}=F\cdot\boldsymbol{\alpha}_{b,v}, \qquad \frac{1}{F}\sum_{j=0}^{F-1}w_{b,v}[j]=1,
\label{eq:meanpreserve}
\end{equation}

under which uniform attention corresponds to $\mathbf{w}\equiv\mathbf{1}$, the identity. The model therefore begins in a neutral state and learns to \emph{amplify or suppress} individual periodicities relative to that state, in the same spirit as the zero-initialized fusion gate of Section~\ref{sec:spatial}. The constraint normalizes the \emph{average gain} across bins and not the mean of the series: the zero-frequency bin is scaled by $w_{b,v}[0]=F\alpha_{b,v}[0]$, so the window mean is preserved only when the attention places exactly uniform mass on that bin. The filtered series is obtained by applying the gain and inverting the transform:

\begin{align}
\tilde{S}_{b,v}[j]&=w_{b,v}[j]\,S_{b,v}[j],\label{eq:fgain}\\
\tilde{\mathbf{s}}_{b,v}&=\operatorname{irFFT}\big(\tilde{S}_{b,v};\,T\big).
\label{eq:ifft}
\end{align}

Since real and imaginary parts are scaled identically, this modulates the \emph{magnitude} of each frequency component while leaving its \emph{phase} unchanged.

The presented mechanism does not produce an attention map that is carried forward: its \emph{only} output is the reweighted series $\tilde{\mathbf{s}}_{b,v}\in\mathbb{R}^{T}$, which substitutes for the raw measurement channel in the input representation of Section~\ref{sec:embedding}. Writing $\tilde{s}_{b,v}[t]$ for its value at step $t$, the measurement fed to the model is

\begin{equation}
\breve{x}^{\mathrm{val}}_{b,t,v}=
\begin{cases}
\tilde{s}_{b,v}[t], & \text{with frequency-bin attention (SAGT-E/F)},\\
\tilde{x}^{\mathrm{val}}_{b,t,v}, & \text{without it (SAGT-D)},
\end{cases}
\label{eq:filtered-channel}
\end{equation}

where $\tilde{x}^{\mathrm{val}}$ is the standardized measurement of \eqref{eq:zscore}. The calendar channels are never filtered. Consequently, the frequency weights influence the forecast solely through the value of $\breve{x}^{\mathrm{val}}$; they are not applied to the temporal attention of Section~\ref{sec:temporal} nor to the spatial operator of Section~\ref{sec:spatial}, both of which act on the embedded tokens.

The distribution $\boldsymbol{\alpha}_{b,v}$ itself is retained at inference purely for analysis: it is a per-sensor profile of which periodicities the model amplifies and plays no further role in the computation.

\subsection{Input Representation}
\label{sec:embedding}

Each triple $(b,t,v)$ is mapped to a token $\mathbf{z}^{(0)}_{b,t,v}\in\mathbb{R}^{d}$ formed by concatenating four complementary components,

\begin{equation} \mathbf{z}^{(0)}_{b,t,v}=\big[\,\mathbf{f}\;\|\;\mathbf{p}^{\mathrm{tod}}\;\|\;\mathbf{p}^{\mathrm{dow}}\;\|\;\mathbf{a}\,\big],
\label{eq:embed}
\end{equation}

whose terms are, respectively, 
\begin{align}
\mathbf{f}&=\mathbf{W}_{\mathrm{in}}\breve{\mathbf{x}}_{b,t,v}+\mathbf{b}_{\mathrm{in}}\in\mathbb{R}^{24},\label{eq:emb-f}\\
\mathbf{p}^{\mathrm{tod}}&=\mathbf{E}^{\mathrm{tod}}\big[\lfloor x^{\mathrm{tod}}_{b,t,v}P\rfloor\big]\in\mathbb{R}^{24},\label{eq:emb-tod}\\
\mathbf{p}^{\mathrm{dow}}&=\mathbf{E}^{\mathrm{dow}}\big[x^{\mathrm{dow}}_{b,t,v}\big]\in\mathbb{R}^{24},\label{eq:emb-dow}\\
\mathbf{a}&=\mathbf{E}^{a}_{t,v}\in\mathbb{R}^{80},\label{eq:emb-a}
\end{align}

giving $d=24+24+24+80=152$. The vector $\breve{\mathbf{x}}_{b,t,v}$ in~\eqref{eq:emb-f} is the full input channel triple at that position, with the measurement channel taken \emph{after} frequency-bin filtering,

\begin{equation}
\breve{\mathbf{x}}_{b,t,v}=\big[\;\breve{x}^{\mathrm{val}}_{b,t,v},\; x^{\mathrm{tod}}_{b,t,v},\; x^{\mathrm{dow}}_{b,t,v}\;\big]^{\!\top} \in\mathbb{R}^{C},
\label{eq:xtilde}
\end{equation}

with $\breve{x}^{\mathrm{val}}$ given by~\eqref{eq:filtered-channel}. This is the single point at which the output of Section~\ref{sec:freqattn} enters the network. The projection $\mathbf{W}_{\mathrm{in}}\in\mathbb{R}^{24\times C}$ therefore acts on all three channels jointly, so the calendar covariates are represented twice and in two different ways: as raw scalars inside this projection, which preserves their ordinal structure, and as learnable embeddings $\mathbf{E}^{\mathrm{tod}}\in\mathbb{R}^{P\times24}$ and $\mathbf{E}^{\mathrm{dow}}\in\mathbb{R}^{7\times24}$ indexed by the discretized covariate, which allows each time slot and weekday to acquire an arbitrary, non-ordinal signature. The latter is what lets the model represent recurring daily and weekly regimes that are not smooth functions of clock time.

The fourth term is the \emph{spatio-temporal adaptive embedding} $\mathbf{E}^{a}\in\mathbb{R}^{T\times N\times80}$: a free parameter tensor, initialized with Xavier-uniform sampling and indexed by (position-in-window, sensor). It is not derived from any input and is trained end-to-end, allowing each sensor to acquire a position-dependent signature without requiring a predefined graph. This component is adopted from STAEformer~\cite{liu2023staeformer}, where it was shown to contribute substantially to that model's accuracy.

Collecting these tokens over the batch gives the \emph{token tensor}

\begin{equation}
\mathcal{Z}^{(0)}\in\mathbb{R}^{B\times T\times N\times d},\qquad \mathcal{Z}^{(0)}[b,t,v,:]=\mathbf{z}^{(0)}_{b,t,v},
\label{eq:ztensor}
\end{equation}

which is the object every subsequent layer transforms: block $\ell$ maps $\mathcal{Z}^{(\ell-1)}\mapsto\mathcal{Z}^{(\ell)}$ without changing its shape, so $\mathcal{Z}^{(\ell)}$ is the same collection of tokens after $\ell$ rounds of refinement. It stands in the same relation to the network as $\mathcal{X}$ in~\eqref{eq:input} does to the raw data, with the channel axis $C$ replaced by the model width $d$.

The temporal and spatial layers read this one tensor along \emph{different} axes, and we name the two views once here so that later sections are unambiguous:

\begin{equation}
\underbrace{\mathbf{Z}^{(\ell)}_{b,:,v}\in\mathbb{R}^{T\times d}}_{\text{one sensor, all }T\text{ steps}}, \qquad \underbrace{\mathbf{Z}^{(\ell)}_{b,t,:}\in\mathbb{R}^{N\times d}}_{\text{one step, all }N\text{ sensors}},
\label{eq:slices}
\end{equation}

where a colon marks the axis that is retained. The temporal layer of Section~\ref{sec:temporal} operates on the first (\emph{temporal}) slice and the hybrid spatial layer of Section~\ref{sec:spatial} on the second (\emph{spatial}) slice; both are views of the same tensor, so no data is duplicated and the two layers simply mix along complementary directions. Where the block index is immaterial, we drop the superscript $(\ell)$.

\subsection{Temporal Transformer Layer}
\label{sec:temporal}

Temporal dependencies are modeled with multi-head self-attention along the time axis, applied independently per sensor. Concretely, it acts on the \emph{temporal} slice $\mathbf{Z}_{b,:,v}\in\mathbb{R}^{T\times d}$ of~\eqref{eq:slices}: the $T$ tokens belonging to sensor $v$ in sample $b$, stacked in chronological order. The same weights are applied to each of the $BN$ such slices, so the sensor index never enters the computation, and no information crosses sensors here. Since $b$ and $v$ are spectators throughout this subsection, we abbreviate $\mathbf{Z}\equiv\mathbf{Z}_{b,:,v}$. The queries, keys and values are $\mathbf{Q}=\mathbf{Z}\mathbf{W}^{Q}$,$\mathbf{K}=\mathbf{Z}\mathbf{W}^{K}$ and $\mathbf{V}=\mathbf{Z}\mathbf{W}^{V}$ with $\mathbf{W}^{Q},\mathbf{W}^{K},\mathbf{W}^{V}\in\mathbb{R}^{d\times d}$, split into $h$ heads of width $d_h=d/h$. Head $i$ computes

\begin{equation}
\mathbf{O}^{(i)}=\operatorname{softmax}\!\left(\frac{\mathbf{Q}^{(i)}\mathbf{K}^{(i)\top}}{\sqrt{d_h}}\right)\mathbf{V}^{(i)},
\label{eq:mha}
\end{equation}

and the heads are concatenated and projected, $\operatorname{MHA}(\mathbf{Z})=[\mathbf{O}^{(1)}\|\cdots\|\mathbf{O}^{(h)}]\mathbf{W}^{O}$. Scaling by $\sqrt{d_h}$ keeps the logits numerically stable; no causal mask is applied, since the entire input window is observed. The layer uses the standard residual and normalization structure,
\begin{align}
\mathbf{Z}'&=\operatorname{LN}\!\big(\mathbf{Z}+\operatorname{Drop}(\operatorname{MHA}(\mathbf{Z}))\big),\label{eq:tblock1}\\
\mathbf{Z}''&=\operatorname{LN}\!\big(\mathbf{Z}'+\operatorname{Drop}(\operatorname{FFN}(\mathbf{Z}'))\big),\label{eq:tblock2}
\end{align}
with $\operatorname{FFN}(\mathbf{u})=\mathbf{W}_b\,\rho(\mathbf{W}_a\mathbf{u})$, $\mathbf{W}_a\in\mathbb{R}^{d_f\times d}$, $\mathbf{W}_b\in\mathbb{R}^{d\times d_f}$, $\operatorname{LN}$ layer normalization and $\operatorname{Drop}$ dropout of rate $p$. Since $T=12$ is small, this pathway costs $\mathcal{O}(BNT^2d)$.

\subsection{Hybrid Spectral Spatial Layer}
\label{sec:spatial}

This layer replaces dense inter-sensor attention with two inexpensive, complementary branches: a \emph{global} low-rank spectral operator and a \emph{local} road-graph diffusion, which are then fused. Where the temporal layer sliced the token tensor along time, this layer takes the complementary \emph{spatial} slice of~\eqref{eq:slices}, together with its rows:

\begin{equation}
\mathbf{Z}_{b,t}\equiv\mathbf{Z}_{b,t,:}\in\mathbb{R}^{N\times d},\qquad \mathbf{z}_{b,t,v}\equiv\mathbf{Z}_{b,t}[v,:]\in\mathbb{R}^{d}.
\label{eq:spatial-slice}
\end{equation}

That is, $\mathbf{Z}_{b,t}$ gathers the tokens of \emph{all} $N$ sensors at the single step $t$ of sample $b$, and $\mathbf{z}_{b,t,v}$ is its $v$-th row, the token of one sensor, which is precisely the vector the preceding temporal layer has just updated. The layer is applied independently to each of the pairs $(b,t)$, so sensors interact while time steps do not; temporal and spatial mixing are kept in separate, alternating layers rather than in one joint operator.

\subsubsection{Directed Low-Rank Spectral Branch}

Instead of computing the Laplacian eigenbasis, we \emph{learn} two bases. Two free parameter matrices $\mathbf{W}_q,\mathbf{W}_k\in\mathbb{R}^{N\times k}$ are re-parameterized through the reduced QR factorization,

\begin{equation}
\mathbf{U}_q=\operatorname{QR}(\mathbf{W}_q).\mathbf{Q},\quad \mathbf{U}_k=\operatorname{QR}(\mathbf{W}_k).\mathbf{Q},
\label{eq:qr}
\end{equation}

so that $\mathbf{U}_q^{\!\top}\mathbf{U}_q=\mathbf{U}_k^{\!\top}\mathbf{U}_k=\mathbf{I}_k$. The factorization is differentiable; hence, orthonormality is enforced exactly at every optimization step rather than encouraged by a penalty. Each of the $k$ columns of $\mathbf{U}$ can be read as a spatial mode: a weighting pattern over all $N$ sensors that behaves coherently, analogous to a graph-Fourier mode but obtained from data.

The filter coefficients are made \emph{input-dependent}. Pooling tokens over sensors yields a context vector describing the current network-wide traffic state, from which a two-layer perceptron predicts one coefficient per mode:

\begin{align}
\mathbf{c}_{b,t}&=\frac{1}{N}\sum_{v=1}^{N}\mathbf{z}_{b,t,v}\in\mathbb{R}^{d},\label{eq:ctx}\\
\boldsymbol{\theta}_{b,t}&=\mathbf{W}_4\,\rho\big(\mathbf{W}_3\mathbf{c}_{b,t}+\mathbf{b}_3\big)+\mathbf{b}_4\in\mathbb{R}^{k},\label{eq:theta}
\end{align}

with $\mathbf{W}_3\in\mathbb{R}^{64\times d}$ and $\mathbf{W}_4\in\mathbb{R}^{k\times64}$. Two properties are deliberate. First, $\boldsymbol{\theta}$ is produced \emph{per time step}, so spatial mixing may differ between, say, the onset and the dissipation of congestion within the same window. Second, no squashing function is applied, so coefficients may be \emph{negative}: the operator can form contrasts between regions rather than only smoothing them, which a non-negative filter cannot express.

Mixing proceeds by analysis, filtering, and synthesis:

\begin{align}
\hat{\mathbf{Z}}_{b,t}&=\mathbf{U}_k^{\!\top}\mathbf{Z}_{b,t}\in\mathbb{R}^{k\times d},\label{eq:analysis}\\
\hat{\mathbf{Z}}_{b,t}[j,:]&\leftarrow\theta_{b,t}[j]\,\hat{\mathbf{Z}}_{b,t}[j,:],\label{eq:filter}\\
\mathbf{G}^{\mathrm{glo}}_{b,t}&=\mathbf{U}_q\hat{\mathbf{Z}}_{b,t},\label{eq:synthesis}
\end{align}

which is equivalent to applying a single dynamic operator,

\begin{equation}
\mathbf{G}^{\mathrm{glo}}_{b,t}=\mathbf{M}_{b,t}\mathbf{Z}_{b,t},\quad \mathbf{M}_{b,t}=\mathbf{U}_q\operatorname{diag}(\boldsymbol{\theta}_{b,t})\mathbf{U}_k^{\!\top},
\label{eq:operator}
\end{equation}

of rank at most $k$. Crucially, $\mathbf{M}_{b,t}$ is never formed explicitly: computation always passes through the $k$-dimensional bottleneck, costing $\mathcal{O}(Nkd)$ instead of $\mathcal{O}(N^2d)$, and no $N\times N$ tensor is stored. Because $\mathbf{U}_q\neq\mathbf{U}_k$ the operator is \emph{asymmetric}, representing directed influence; because $\boldsymbol{\theta}$ is signed it is \emph{indefinite}. Comparing \eqref{eq:operator} with the classical form \eqref{eq:classical} makes three substitutions explicit: the fixed eigenbasis becomes a learned truncated basis, the static filter becomes an input-conditioned one, and the symmetric sandwich becomes a directed pair.

\subsubsection{Sparse Local Diffusion Branch}

Low-rank global mixing captures network-wide structure but, by construction, cannot represent strictly local propagation, even though congestion physically travels along individual road segments. We therefore add a sparse branch operating directly on the road graph. Two random-walk normalizations give forward and backward transition operators,

\begin{equation}
\mathbf{A}_{\rightarrow}=\mathbf{D}_{O}^{-1}\mathbf{A},\qquad \mathbf{A}_{\leftarrow}=\mathbf{D}_{I}^{-1}\mathbf{A}^{\!\top},
\label{eq:supports}
\end{equation}

where $\mathbf{D}_{O}=\operatorname{diag}(\mathbf{A}\mathbf{1})$ and $\mathbf{D}_{I}=\operatorname{diag}(\mathbf{A}^{\!\top}\mathbf{1})$ are the out- and in-degree matrices, with zero-degree rows set to zero. The branch is a one-hop directed diffusion,

\begin{equation}
\mathbf{G}^{\mathrm{loc}}_{b,t}=\mathbf{A}_{\rightarrow}\mathbf{Z}_{b,t}\mathbf{W}_{\rightarrow} +\mathbf{A}_{\leftarrow}\mathbf{Z}_{b,t}\mathbf{W}_{\leftarrow},
\label{eq:local}
\end{equation}

with $\mathbf{W}_{\rightarrow},\mathbf{W}_{\leftarrow}\in\mathbb{R}^{d\times d}$, implemented with sparse products so that its cost is $\mathcal{O}(|E|d)$. Since road networks are sparse ($|E|\ll N^2$), this branch is inexpensive. We keep it deliberately simple: it is standard machinery whose role is to supply the locality that the low-rank branch cannot.

\subsubsection{Zero-Initialized Fusion}

The two branches are concatenated per sensor and projected back to the model width,

\begin{equation}
\mathbf{G}^{\mathrm{fus}}_{b,t,v}=\mathbf{W}_{f}\big[\mathbf{G}^{\mathrm{loc}}_{b,t,v}\,\|\, \mathbf{G}^{\mathrm{glo}}_{b,t,v}\big]+\mathbf{b}_{f},
\label{eq:fuse}
\end{equation}

where $\mathbf{G}^{\mathrm{loc}}_{b,t,v}$ and $\mathbf{G}^{\mathrm{glo}}_{b,t,v}$ are row $v$ of \eqref{eq:local} and \eqref{eq:synthesis} respectively, the two branches' contributions to the same sensor, and $\mathbf{W}_{f}\in\mathbb{R}^{d\times2d}$. Stacking these rows over sensors gives $\mathbf{G}^{\mathrm{fus}}_{b,t}\in\mathbb{R}^{N\times d}$, which enters the residual and feed-forward structure of \eqref{eq:tblock1}--\eqref{eq:tblock2}:

\begin{align}
\mathbf{Z}'_{b,t}&=\operatorname{LN}\!\big(\mathbf{Z}_{b,t}+\operatorname{Drop}(\mathbf{G}^{\mathrm{fus}}_{b,t})\big),\label{eq:sblock1}\\ \mathbf{Z}''_{b,t}&=\operatorname{LN}\!\big(\mathbf{Z}'_{b,t}+\operatorname{Drop}(\operatorname{FFN}(\mathbf{Z}'_{b,t}))\big).\label{eq:sblock2}
\end{align}

We initialize $\mathbf{W}_{f}=\mathbf{0}$ and $\mathbf{b}_{f}=\mathbf{0}$. The entire spatial layer, therefore, starts as the identity map, and training grows each branch only until it reduces the loss, thereby stabilizing early optimization while the learned bases are still random.

One implementation detail is worth mentioning: the fusion projection must be the \emph{only} zero-initialized gate. If the output layer of the filter network in \eqref{eq:theta} were also zeroed, the two would gate each other: the gradient with respect to $\boldsymbol{\theta}$ is proportional to the global block of $\mathbf{W}_{f}$, while the gradient with respect to that block is proportional to $\mathbf{G}^{\mathrm{glo}}$. Both being zero is a stationary point, and the global branch would remain permanently inactive. We verified empirically that the single-gate configuration delivers non-zero gradients to $\mathbf{W}_q$, $\mathbf{W}_k$, and the local branch from the first optimization steps.

The complete encoder stacks $L$ blocks, each a temporal layer (Section~\ref{sec:temporal}) followed by a hybrid spatial layer, with independent parameters per block. Block $\ell$ receives $\mathcal{Z}^{(\ell-1)}$, applies \eqref{eq:tblock1}--\eqref{eq:tblock2} to every temporal slice and then \eqref{eq:sblock1}--\eqref{eq:sblock2} to every spatial slice, and writes the result back as $\mathcal{Z}^{(\ell)}$. The shape $B\times T\times N\times d$ is preserved throughout, so the two layers alternate over the same tensor rather than passing reshaped intermediates between them.

\subsection{Model Variants and the Single-Basis Predecessor}
\label{sec:variants}

SAGT was built incrementally, and the intermediate variants are reported in Section~\ref{sec:results} so that each design decision can be attributed. Since two of them use an \emph{earlier} spatial layer rather than a reduced version of the one above, we define that layer explicitly here.

The predecessor is a \emph{single-basis} spectral layer. It uses one learned orthonormal basis $\mathbf{U}=\operatorname{QR}(\mathbf{W}_u).\mathbf{Q}\in\mathbb{R}^{N\times k}$ on both sides, and its filter is produced once per input window from the context pooled over \emph{both} sensors and time steps:

\begin{align}
\mathbf{c}_{b}&=\operatorname{vec}\big(\big[\mathbf{c}_{b,0},\dots,\mathbf{c}_{b,T-1}\big]\big)\in\mathbb{R}^{Td},\label{eq:sb-ctx}\\
\boldsymbol{\varphi}_{b}&=\varsigma\big(\mathbf{W}_6\,\rho(\mathbf{W}_5\mathbf{c}_{b}+\mathbf{b}_5)+\mathbf{b}_6\big)\in(0,1)^{k},\label{eq:sb-phi}\\
\mathbf{M}^{\mathrm{sb}}_{b}&=\mathbf{U}\operatorname{diag}(\boldsymbol{\varphi}_{b})\mathbf{U}^{\!\top},\label{eq:sb-op}
\end{align}

where $\mathbf{c}_{b,t}$ is the sensor-pooled context of \eqref{eq:ctx}, $\mathbf{W}_5\in\mathbb{R}^{64\times Td}$, $\mathbf{W}_6\in\mathbb{R}^{k\times64}$, and $\varsigma$ is the logistic sigmoid. The mixed signal 
$\mathbf{M}^{\mathrm{sb}}_{b}\mathbf{Z}_{b,t}$ is passed through a linear projection and then the same residual and feed-forward structure as \eqref{eq:sblock1}--\eqref{eq:sblock2}; there is no local branch and no zero-initialized gate.

The contrast with Section~\ref{sec:spatial} is exactly the contrast drawn in Section~\ref{sec:rationale}, and it is instructive that the predecessor sits on the wrong side of all three points. As a single basis is used on both sides, $\mathbf{M}^{\mathrm{sb}}_{b}$ is \emph{symmetric} and cannot express directed influence. Since $\varsigma$ bounds the coefficients to $(0,1)$, it is \emph{positive semi-definite} and can only attenuate spatial modes, never form contrasts between regions. And because the pooling in \eqref{eq:sb-ctx} collapses the time axis, the operator carries no $t$ index: a single filter is applied to the whole window, so mixing cannot differ between the onset and the dissipation of congestion. The hybrid layer of Section~\ref{sec:spatial} removes all three restrictions and adds
the sparse local branch. The three reported variants are then described in Table~\ref{tab:variants}.

\begin{table}[!t]
\renewcommand{\arraystretch}{1.15}
\caption{Reported Variants. All Share the Temporal Backbone of Section~\ref{sec:temporal}, the Embedding of Section~\ref{sec:embedding} and the Head of Section~\ref{sec:head}}
\label{tab:variants}
\centering
\footnotesize
\begin{tabular}{lll}
\toprule
Variant & Spatial layer & Freq.\ attention \\
\midrule
SAGT-D & single-basis, \eqref{eq:sb-op}   & no  \\
SAGT-E & single-basis, \eqref{eq:sb-op}   & yes \\
\textbf{SAGT-F} (full) & hybrid, Sec.~\ref{sec:spatial} & yes \\
\bottomrule
\end{tabular}
\end{table}

SAGT-F is the complete model; unqualified references to SAGT denote it. The three variants form a chain in which each step adds one component: D$\to$E introduces frequency-bin attention and E$\to$F replaces the single-basis operator with the hybrid one, so the two contributions can be read independently from Table~\ref{tab:ladder}. Removing frequency-bin attention from SAGT-F, rather than adding it to SAGT-D, gives the complementary comparison; that configuration is not a named variant and is reported as an ablation in Section~\ref{sec:ablation}.

\subsection{Prediction Head and Objective}
\label{sec:head}

After the final block, each sensor's $T$ processed tokens are flattened and mapped to the forecast horizon by a single linear layer,

\begin{equation}
\hat{\mathbf{y}}_{b,:,v}=\mathbf{W}_{\mathrm{out}}\operatorname{vec}\!\big(\mathbf{Z}^{(L)}_{b,:,v}\big), \quad \mathbf{W}_{\mathrm{out}}\in\mathbb{R}^{H\times Td},
\label{eq:head}
\end{equation}

producing all $H$ steps simultaneously and thereby avoiding the error accumulation of recursive decoding.

Traffic datasets contain missing readings encoded as zeros, which must be excluded from the objective. With indicator $m=1[y\neq0]$ and its mean-normalized form $\bar{m}=m/\operatorname{mean}(m)$, the training loss is the masked Huber criterion

\begin{equation}
\ell_\delta(e)=
\begin{cases}
\tfrac{1}{2}e^{2}, & |e|\le\delta,\\
\delta\big(|e|-\tfrac{\delta}{2}\big), & |e|>\delta,
\end{cases}
\label{eq:huber}
\end{equation}

\begin{equation}
\mathcal{L}=\operatorname{mean}\!\big(\bar{m}\odot\ell_\delta(\hat{y}-y)\big),
\label{eq:loss}
\end{equation}

with $\delta=1$, $e=\hat{y}-y$ and $\odot$ the elementwise product. The Huber form behaves quadratically for small residuals, aiding optimization, and linearly for large ones, limiting the influence of rare outliers. On the three speed datasets (METR-LA, PEMS-BAY and EXPY-TKY), we instead use the masked mean absolute error, $\operatorname{mean}(\bar{m}\odot|\hat{y}-y|)$, matching the convention established in that literature. Baselines are trained to match their reference implementations.

\subsection{Complexity Analysis}
\label{sec:complexity}

For one spatial layer and one input window, the global branch costs $\mathcal{O}(TNkd)$ for the projections plus $\mathcal{O}(Nk^2)$ for the QR factorization; the local branch costs $\mathcal{O}(T|E|d)$ plus $\mathcal{O}(TNd^2)$ for channel mixing; and fusion costs $\mathcal{O}(TNd^2)$. All terms are \emph{linear} in $N$. The comparable attention-based spatial layer requires $\mathcal{O}(TN^2d)$ time and, more restrictively, materializes an $N\times N$ attention map per head at every time step. The practical consequence is a memory profile that grows far more slowly with network size.

\section{Experiments and Results}
\label{sec:results}

\subsection{Datasets}
\label{sec:datasets}

We evaluate on six public traffic benchmark datasets spanning two sensing modalities and almost an order of magnitude in network size (Table~\ref{tab:datasets}). PEMS04, PEMS07 and PEMS08~\cite{guo2019attention,song2020stsgcn} record traffic \emph{flow}; METR-LA and PEMS-BAY~\cite{li2017diffusion}, and EXPY-TKY~\cite{jiang2023spatio}, record \emph{speed}. EXPY-TKY, with $1843$ sensors, is included specifically to probe scalability: it is larger than the benchmarks on which the recent transformer baselines compared here are typically validated.

All datasets use $T=H=12$. Following standard practice we split PEMS04/07/08 chronologically at $6\!:\!2\!:\!2$ and METR-LA/PEMS-BAY/EXPY-TKY at $7\!:\!1\!:\!2$. Every model consumes \emph{identical} tensors produced by one shared data pipeline, and every metric is computed by one shared implementation, so no difference in preprocessing or evaluation can contribute to the comparison.

\begin{table}[!t]
\renewcommand{\arraystretch}{1.15}
\caption{Benchmark Datasets. $P$ Is the Number of Time Steps per Day}
\label{tab:datasets}
\centering
\footnotesize
\begin{tabular}{lccccc}
\toprule
Dataset & $N$ & Steps & Interval & $P$ & Signal \\
\midrule
PEMS08   & 170  & 17\,856 & 5\,min  & 288 & flow \\
METR-LA  & 207  & 34\,272 & 5\,min  & 288 & speed \\
PEMS04   & 307  & 16\,992 & 5\,min  & 288 & flow \\
PEMS-BAY & 325  & 52\,116 & 5\,min  & 288 & speed \\
PEMS07   & 883  & 28\,224 & 5\,min  & 288 & flow \\
EXPY-TKY & 1843 & 13\,248 & 10\,min & 144 & speed \\
\bottomrule
\end{tabular}
\end{table}

\subsection{Baselines and Protocol}
\label{sec:baselines}

We compare SAGT with eight representative models covering the main design families: STGCN~\cite{yu2017spatio} and DCRNN~\cite{li2017diffusion} (early spatio-temporal graph networks), AGCRN~\cite{bai2020agcrn} (adaptive adjacency), StemGNN~\cite{cao2020spectral} (spectral, the closest prior work to ours), and the recent transformer generation: STAEformer~\cite{liu2023staeformer}, PDFormer~\cite{jiang2023pdformer}, TESTAM~\cite{lee2024testam} and MegaCRN~\cite{jiang2023spatio}.

We additionally report GC-MoE~\cite{gcmoe}, which is not a single end-to-end model but a learned router over three frozen, independently pretrained forecasters. Its capacity is the union of several networks, and it is trained under a different protocol, so we show it in every table for completeness, but exclude it from the ranking. Section~\ref{sec:composite} compares against it directly.

Errors are reported as mean absolute error (MAE), root mean squared error (RMSE), and mean absolute percentage error (MAPE), each masked to exclude missing readings and averaged over the $12$ horizons; we report the same metrics at horizons $3$, $6$, and $12$. Every entry is the mean $\pm$ sample standard deviation over three training runs that differ only in the random seed. For GC-MoE, only the router is retrained per seed; its frozen experts are the same STGCN, AGCRN, and Graph WaveNet checkpoints across all seeds. Runs use Adam~\cite{kingma2014adam} with gradient-norm clipping at $5$, a multi-step schedule decaying by $0.1$ at epochs $\{25,45,65\}$, at most $200$ epochs, and early stopping with patience $25$ on validation loss.

What we deliberately do not unify is the learning rate. Imposing a single value across architectures would penalize those whose published configuration assumes another value, inflating our own results rather than controlling for anything; each baseline, therefore, keeps the rate its authors report. 

Batch size is handled on the same principle as the learning rate, and for the same reason. Each model uses the batch size its authors report, reduced only where it does not fit in the $16$\,GB of our GPUs. Weight decay is $5\times10^{-4}$ on the flow benchmarks and $3\times10^{-4}$ on the speed benchmarks. All SAGT variants use a spectral rank $k=32$, held fixed across all datasets and variants, rather than being tuned per benchmark.

\subsubsection{Reproduction Fidelity}

Reproducing published baselines faithfully requires care, and we document two cases in detail for transparency.

For \emph{TESTAM}~\cite{lee2024testam} model, our re-implementation matched the authors' released code. During these runs, we encountered a latent numerical instability in the released code, worth documenting as it affects the reported score. The routing-label computation normalizes the per-expert losses by their sum after zeroing the worst expert; that denominator is exactly zero whenever no expert other than the worst incurs a loss, and the division is unguarded. A second unguarded expression takes the logarithm of the gate, which is a softmax and may underflow to zero, so that the routing term evaluates $0\times(-\infty)$. Either path introduces NaN into the objective, and because the masked error converts NaN to zero, the corrupted state can subsequently be recorded as the best validation score. To address the issue, we added clamped denominators and a NaN guard, leaving the method itself unchanged.

For the \emph{StemGNN}~\cite{cao2020spectral} model, the published figures were obtained under a protocol that differs structurally from a unified benchmark: prediction at horizon $3$ rather than a $12$-step average, the speed channel of the PEMS04 dataset rather than flow, a $228$-node variant of the PEMS07 dataset, a mean-squared objective in normalized space, and unmasked metrics. Its published values are therefore not directly comparable with the remaining entries. We report them under our unified protocol. Comparable deviations for this model are documented by standardized benchmark suites such as BasicTS~\cite{shao2024exploring} and LibCity~\cite{wang2021libcity}.

Within a fixed hardware and software environment, our runs are deterministic: repeating a completed SAGT run from its logged configuration exactly reproduces the parameter count and the per-epoch training and validation losses to four decimal places. We use this as a routine check to ensure that the reported number and the configuration printed beside it were, in fact, produced together.

\subsection{Main Results}
\label{sec:main}

Table~\ref{tab:main} reports the averaged MAE, RMSE, and MAPE for all models on all six benchmarks, each as the mean $\pm$ standard deviation over three seeds. SAGT-F attains the best MAE, RMSE and MAPE of any single end-to-end model on PEMS08 ($14.39\pm0.11$~MAE, ahead of TESTAM by $0.25$ and STAEformer by $0.40$). On PEMS04, it is third in MAE ($18.89\pm0.13$), behind PDFormer ($18.23\pm0.06$) and TESTAM ($18.73\pm0.03$), although it has the second-lowest RMSE. On PEMS07, it is third in MAE ($20.57\pm0.05$), $0.52$ behind TESTAM and $0.10$ behind STAEformer, and second in MAPE. On EXPY-TKY, it is $0.12$ behind TESTAM and $0.06$ behind STAEformer, again with the second-lowest MAPE, and on the two speed benchmarks, it trails the leader by $0.13$ (METR-LA) and $0.10$ (PEMS-BAY). TESTAM is the strongest baseline overall: it leads on PEMS-BAY, PEMS07, and EXPY-TKY, and shares the lead on METR-LA with STAEformer.

Seed variation is far from uniform across the table. Most entries have a standard deviation below $0.2$~MAE, but several baselines are markedly unstable under our protocol: PDFormer on PEMS08 ($15.27\pm0.82$~MAE, $14.94\pm3.37\%$ MAPE), MegaCRN on PEMS07 ($28.00\pm4.05$ MAE) and, above all, StemGNN on PEMS07 ($33.47\pm7.32$ MAE), where two of the three seeds stall on a validation plateau and stop early. A single noise figure for the whole table would therefore be misleading, and differences smaller than the standard deviations involved should be read as ties.

Read as a whole, SAGT-F wins one benchmark outright, is within $0.13$~MAE of the leader on three of the remaining five, and trails by $0.5$--$0.7$~MAE on PEMS04 and PEMS07. That is a competitive but not dominant accuracy profile. The argument of this paper rests on the cost at which such a result is obtained, which Section~\ref{sec:efficiency} quantifies.

\begin{table*}[!t]
\renewcommand{\arraystretch}{1.3}
\caption{Averaged MAE, RMSE, and MAPE Over 12 Horizons on Six Benchmarks. Ranking (\textbf{Bold} Best, \underline{Underlined} Second) Is Computed per Metric Over Single End-to-End Models Only. $^{\dagger}$GC-MoE Composes Three Frozen Pretrained Experts and Is Shown for Completeness but Not Ranked (Sec.~\ref{sec:composite}); Its Cost Accounting Appears in Table~\ref{tab:eff}.}
\label{tab:main}
\centering
\scriptsize
\begin{tabular}{lccccccccc}
\toprule
\multirow{2}{*}{Model} & \multicolumn{3}{c}{PEMS08} & \multicolumn{3}{c}{METR-LA} & \multicolumn{3}{c}{PEMS04} \\
\cmidrule(lr){2-4}\cmidrule(lr){5-7}\cmidrule(lr){8-10} & MAE & RMSE & MAPE & MAE & RMSE & MAPE & MAE & RMSE & MAPE \\
\midrule
STGCN & 16.66$\pm$0.07 & 25.85$\pm$0.06 & 11.15$\pm$0.31\% & 3.13$\pm$0.00 & 6.35$\pm$0.02 & 8.70$\pm$0.07\% & 20.36$\pm$0.13 & 32.16$\pm$0.33 & 14.26$\pm$0.30\% \\
DCRNN & 15.64$\pm$0.06 & 24.66$\pm$0.06 & 10.21$\pm$0.07\% & 3.59$\pm$0.39 & 7.04$\pm$0.56 & 9.39$\pm$0.31\% & 20.71$\pm$0.09 & 32.44$\pm$0.10 & 14.57$\pm$0.14\% \\
AGCRN & 15.86$\pm$0.11 & 25.15$\pm$0.07 & 10.81$\pm$0.53\% & 3.19$\pm$0.01 & 6.45$\pm$0.01 & 8.86$\pm$0.06\% & 19.40$\pm$0.05 & 31.26$\pm$0.15 & 13.44$\pm$0.28\% \\
StemGNN & 17.53$\pm$0.55 & 27.52$\pm$0.60 & 11.37$\pm$0.46\% & 3.71$\pm$0.28 & 7.32$\pm$0.43 & 10.77$\pm$1.00\% & 22.96$\pm$0.51 & 35.59$\pm$0.59 & 15.94$\pm$0.31\% \\
MegaCRN & 15.42$\pm$0.30 & 24.39$\pm$0.37 & 9.95$\pm$0.30\% & 4.19$\pm$0.24 & 10.12$\pm$0.65 & 10.35$\pm$0.49\% & 19.91$\pm$1.15 & 31.44$\pm$0.90 & 14.64$\pm$1.98\% \\
PDFormer & 15.27$\pm$0.82 & 23.99$\pm$0.25 & 14.94$\pm$3.37\% & 3.09$\pm$0.04 & 6.28$\pm$0.01 & 8.64$\pm$0.16\% & \textbf{18.23$\pm$0.06} & \textbf{29.85$\pm$0.08} & \textbf{12.45$\pm$0.13\%} \\
STAEformer & 14.79$\pm$0.11 & \underline{23.97$\pm$0.14} & 9.61$\pm$0.14\% & \textbf{2.99$\pm$0.05} & \underline{6.12$\pm$0.17} & \underline{8.32$\pm$0.22\%} & 19.13$\pm$0.16 & 30.67$\pm$0.22 & 13.39$\pm$0.17\% \\
TESTAM & \underline{14.64$\pm$0.15} & 24.04$\pm$0.22 & \underline{9.50$\pm$0.05\%} & \textbf{2.99$\pm$0.05} & \textbf{6.09$\pm$0.13} & \textbf{8.21$\pm$0.21\%} & \underline{18.73$\pm$0.03} & 31.45$\pm$0.18 & \underline{12.62$\pm$0.07\%} \\
\midrule
SAGT-D & 14.65$\pm$0.29 & 24.03$\pm$0.01 & 9.60$\pm$0.18\% & 3.14$\pm$0.01 & 6.34$\pm$0.05 & 8.90$\pm$0.21\% & 19.14$\pm$0.19 & 31.12$\pm$0.30 & 13.38$\pm$0.22\% \\
SAGT-E & 15.00$\pm$0.13 & 24.16$\pm$0.15 & 9.69$\pm$0.18\% & 3.17$\pm$0.04 & 6.50$\pm$0.08 & 8.95$\pm$0.18\% & 19.04$\pm$0.13 & 30.98$\pm$0.31 & 13.10$\pm$0.01\% \\
\textbf{SAGT-F (ours)} & \textbf{14.39$\pm$0.11} & \textbf{23.62$\pm$0.01} & \textbf{9.36$\pm$0.11\%} & 3.12$\pm$0.01 & 6.36$\pm$0.09 & 8.75$\pm$0.10\% & 18.89$\pm$0.13 & \underline{30.52$\pm$0.33} & 13.14$\pm$0.20\% \\

\midrule
GC-MoE$^{\dagger}$ & 14.42$\pm$0.01 & 23.32$\pm$0.03 & 9.29$\pm$0.00\% & 3.00$\pm$0.00 & 5.99$\pm$0.00 & 8.18$\pm$0.00\% & 18.83$\pm$0.01 & 30.51$\pm$0.01 & 12.83$\pm$0.01\% \\
\midrule[\heavyrulewidth]

\multirow{2}{*}{Model} & \multicolumn{3}{c}{PEMS-BAY} & \multicolumn{3}{c}{PEMS07} & \multicolumn{3}{c}{EXPY-TKY} \\
\cmidrule(lr){2-4}\cmidrule(lr){5-7}\cmidrule(lr){8-10}
 & MAE & RMSE & MAPE & MAE & RMSE & MAPE & MAE & RMSE & MAPE \\
\midrule
STGCN & 1.69$\pm$0.01 & 3.79$\pm$0.04 & 3.85$\pm$0.05\% & 23.54$\pm$0.59 & 36.71$\pm$0.50 & 10.48$\pm$0.70\% & 7.02$\pm$0.04 & 11.28$\pm$0.09 & 32.49$\pm$0.34\% \\
DCRNN & 2.03$\pm$0.10 & 4.49$\pm$0.18 & 4.37$\pm$0.20\% & 30.43$\pm$1.35 & 45.01$\pm$1.75 & 17.29$\pm$2.65\% & 7.80$\pm$0.14 & 12.46$\pm$0.21 & 38.02$\pm$4.07\% \\
AGCRN & 1.62$\pm$0.01 & 3.71$\pm$0.01 & 3.68$\pm$0.01\% & 20.66$\pm$0.04 & 34.44$\pm$0.07 & 8.83$\pm$0.07\% & 6.84$\pm$0.02 & 11.03$\pm$0.04 & 31.58$\pm$0.09\% \\
StemGNN & 1.92$\pm$0.06 & 4.37$\pm$0.16 & 4.53$\pm$0.09\% & 33.47$\pm$7.32 & 49.82$\pm$9.63 & 19.53$\pm$7.34\% & 7.35$\pm$0.07 & 11.68$\pm$0.09 & 34.78$\pm$0.44\% \\
MegaCRN & 1.68$\pm$0.03 & 3.98$\pm$0.04 & 3.87$\pm$0.06\% & 28.00$\pm$4.05 & 42.54$\pm$5.61 & 13.26$\pm$2.21\% & 7.36$\pm$0.02 & 12.11$\pm$0.24 & 31.25$\pm$1.37\% \\
PDFormer & 1.58$\pm$0.01 & 3.62$\pm$0.02 & \underline{3.56$\pm$0.04\%} & 22.96$\pm$0.68 & 35.08$\pm$0.52 & 14.71$\pm$0.69\% & 7.12$\pm$0.03 & 11.32$\pm$0.01 & 33.50$\pm$0.56\% \\
STAEformer & \underline{1.57$\pm$0.02} & \underline{3.59$\pm$0.02} & \underline{3.56$\pm$0.03\%} & \underline{20.47$\pm$0.09} & \underline{33.79$\pm$0.09} & 8.65$\pm$0.08\% & \underline{6.78$\pm$0.03} & \underline{10.96$\pm$0.04} & 31.26$\pm$0.24\% \\
TESTAM & \textbf{1.53$\pm$0.01} & \textbf{3.48$\pm$0.03} & \textbf{3.46$\pm$0.03\%} & \textbf{20.05$\pm$0.14} & \textbf{33.38$\pm$0.33} & \textbf{8.39$\pm$0.08\%} & \textbf{6.72$\pm$0.06} & \textbf{10.91$\pm$0.10} & \textbf{30.85$\pm$0.14\%} \\
\midrule
SAGT-D & 1.64$\pm$0.03 & 3.81$\pm$0.07 & 3.79$\pm$0.10\% & 20.88$\pm$0.12 & 34.65$\pm$0.15 & 8.83$\pm$0.07\% & 6.89$\pm$0.04 & 11.13$\pm$0.05 & 31.19$\pm$0.48\% \\
SAGT-E & 1.62$\pm$0.01 & 3.72$\pm$0.02 & 3.71$\pm$0.02\% & 20.94$\pm$0.08 & 34.69$\pm$0.08 & 8.84$\pm$0.07\% & 6.93$\pm$0.06 & 11.18$\pm$0.12 & 31.26$\pm$0.24\% \\
\textbf{SAGT-F (ours)} & 1.63$\pm$0.03 & 3.77$\pm$0.10 & 3.71$\pm$0.09\% & 20.57$\pm$0.05 & 34.00$\pm$0.17 & \underline{8.63$\pm$0.03\%} & 6.84$\pm$0.01 & 11.02$\pm$0.03 & \underline{31.02$\pm$0.22\%} \\

\midrule
GC-MoE$^{\dagger}$ & 1.57$\pm$0.00 & 3.58$\pm$0.00 & 3.56$\pm$0.00\% & 20.17$\pm$0.01 & 33.63$\pm$0.01 & 8.64$\pm$0.04\% & 6.73$\pm$0.01 & 10.78$\pm$0.01 & 30.94$\pm$0.03\% \\
\bottomrule
\end{tabular}
\end{table*}

Table~\ref{tab:horizon} gives the horizon-resolved breakdown. The same four ranked models are shown on every dataset: the three strongest recent transformer baselines and ours, with GC-MoE added for reference. SAGT-F leads on PEMS08 at every horizon, with a margin over TESTAM that narrows with lead time ($-0.34$, $-0.21$ and $-0.12$~MAE at $h{=}3,6,12$). On PEMS04, it is marginally ahead of TESTAM at $h{=}3$ ($-0.07$) but falls behind at longer horizons ($+0.17$, $+0.58$), while staying ahead of STAEformer at all three. The gap to TESTAM widens monotonically with horizon on PEMS07 ($+0.37$, $+0.54$, $+0.82$), which suggests that its mixture-of-experts routing extrapolates further than our single-operator design on large graphs; on EXPY-TKY, by contrast, the gap narrows with horizon ($+0.17$, $+0.14$, $+0.09$), and SAGT-F has the lowest MAPE at $h{=}12$.

\begin{table*}[!t]
\renewcommand{\arraystretch}{1.1}
\caption{Horizon-Resolved MAE, RMSE, and MAPE at 3, 6, and 12 Steps Ahead. For Consistency, the Same Four Ranked Models Are Shown on Every Dataset: the Three Strongest Recent Transformer Baselines and Ours. \textbf{Bold} Marks the Best per Column Within Each Dataset. $^{\dagger}$GC-MoE Is a Composite Method, Shown for Reference and Not Ranked}
\label{tab:horizon}
\centering
\scriptsize
\begin{tabular}{lccccccccc}
\toprule
\multirow{2}{*}{Model} & \multicolumn{3}{c}{Horizon 3} & \multicolumn{3}{c}{Horizon 6} & \multicolumn{3}{c}{Horizon 12} \\
\cmidrule(lr){2-4}\cmidrule(lr){5-7}\cmidrule(lr){8-10}
 & MAE & RMSE & MAPE & MAE & RMSE & MAPE & MAE & RMSE & MAPE \\
\midrule
\multicolumn{10}{c}{\textit{PEMS08}} \\
\cmidrule{1-10}
PDFormer & 14.50$\pm$0.74 & 22.42$\pm$0.35 & 14.34$\pm$3.33\% & 15.25$\pm$0.85 & 24.01$\pm$0.28 & 14.98$\pm$3.45\% & 16.61$\pm$0.88 & 26.15$\pm$0.32 & 16.26$\pm$3.68\% \\
STAEformer & 13.88$\pm$0.08 & 22.29$\pm$0.11 & 8.94$\pm$0.12\% & 14.75$\pm$0.12 & 23.96$\pm$0.15 & 9.54$\pm$0.12\% & 16.36$\pm$0.13 & 26.50$\pm$0.15 & 10.80$\pm$0.21\% \\
TESTAM & 13.77$\pm$0.12 & 22.40$\pm$0.16 & 8.94$\pm$0.03\% & 14.59$\pm$0.13 & 24.11$\pm$0.21 & 9.44$\pm$0.03\% & 16.01$\pm$0.20 & 26.39$\pm$0.29 & 10.41$\pm$0.12\% \\
\textbf{SAGT-F (ours)} & \textbf{13.43$\pm$0.15} & \textbf{21.86$\pm$0.07} & \textbf{8.75$\pm$0.15\%} & \textbf{14.38$\pm$0.12} & \textbf{23.66$\pm$0.02} & \textbf{9.27$\pm$0.11\%} & \textbf{15.89$\pm$0.06} & \textbf{26.03$\pm$0.06} & \textbf{10.37$\pm$0.07\%} \\
GC-MoE$^{\dagger}$ & 13.50$\pm$0.00 & 21.52$\pm$0.01 & 8.74$\pm$0.01\% & 14.41$\pm$0.01 & 23.30$\pm$0.03 & 9.13$\pm$0.01\% & 15.78$\pm$0.01 & 25.71$\pm$0.03 & 10.15$\pm$0.01\% \\
\midrule
\multicolumn{10}{c}{\textit{METR-LA}} \\
\cmidrule{1-10}
PDFormer & 2.78$\pm$0.03 & 5.39$\pm$0.01 & 7.35$\pm$0.13\% & 3.14$\pm$0.04 & 6.36$\pm$0.02 & 8.82$\pm$0.17\% & 3.52$\pm$0.04 & 7.32$\pm$0.01 & 10.40$\pm$0.21\% \\
STAEformer & 2.72$\pm$0.05 & 5.26$\pm$0.15 & 7.16$\pm$0.21\% & \textbf{3.03$\pm$0.07} & \textbf{6.19$\pm$0.20} & 8.47$\pm$0.28\% & \textbf{3.38$\pm$0.06} & \textbf{7.12$\pm$0.17} & 9.92$\pm$0.20\% \\
TESTAM & \textbf{2.69$\pm$0.05} & \textbf{5.25$\pm$0.15} & \textbf{6.97$\pm$0.20\%} & \textbf{3.03$\pm$0.06} & 6.22$\pm$0.15 & \textbf{8.33$\pm$0.25\%} & 3.43$\pm$0.05 & 7.24$\pm$0.12 & \textbf{9.90$\pm$0.19\%} \\
\textbf{SAGT-F (ours)} & 2.79$\pm$0.01 & 5.35$\pm$0.05 & 7.38$\pm$0.14\% & 3.17$\pm$0.02 & 6.44$\pm$0.10 & 8.93$\pm$0.10\% & 3.57$\pm$0.02 & 7.48$\pm$0.12 & 10.58$\pm$0.08\% \\
GC-MoE$^{\dagger}$ & 2.69$\pm$0.00 & 5.11$\pm$0.00 & 6.92$\pm$0.00\% & 3.05$\pm$0.00 & 6.05$\pm$0.00 & 8.30$\pm$0.00\% & 3.44$\pm$0.00 & 7.03$\pm$0.01 & 9.92$\pm$0.01\% \\
\midrule
\multicolumn{10}{c}{\textit{PEMS04}} \\
\cmidrule{1-10}
PDFormer & \textbf{17.49$\pm$0.03} & \textbf{28.66$\pm$0.03} & \textbf{11.95$\pm$0.18\%} & \textbf{18.22$\pm$0.07} & \textbf{29.90$\pm$0.10} & \textbf{12.40$\pm$0.15\%} & \textbf{19.57$\pm$0.09} & \textbf{31.75$\pm$0.11} & 13.43$\pm$0.18\% \\
STAEformer & 18.20$\pm$0.10 & 29.25$\pm$0.13 & 12.64$\pm$0.06\% & 19.14$\pm$0.16 & 30.69$\pm$0.21 & 13.35$\pm$0.17\% & 20.73$\pm$0.28 & 32.93$\pm$0.36 & 14.70$\pm$0.35\% \\
TESTAM & 18.04$\pm$0.04 & 30.25$\pm$0.28 & 12.21$\pm$0.08\% & 18.73$\pm$0.03 & 31.57$\pm$0.18 & 12.58$\pm$0.08\% & 19.79$\pm$0.06 & 33.15$\pm$0.09 & \textbf{13.32$\pm$0.06\%} \\
\textbf{SAGT-F (ours)} & 17.97$\pm$0.12 & 29.01$\pm$0.23 & 12.67$\pm$0.25\% & 18.90$\pm$0.14 & 30.54$\pm$0.33 & 13.16$\pm$0.24\% & 20.37$\pm$0.13 & 32.78$\pm$0.45 & 14.05$\pm$0.19\% \\
GC-MoE$^{\dagger}$ & 18.01$\pm$0.01 & 29.10$\pm$0.01 & 12.14$\pm$0.00\% & 18.86$\pm$0.01 & 30.61$\pm$0.01 & 12.84$\pm$0.01\% & 20.11$\pm$0.01 & 32.45$\pm$0.01 & 13.86$\pm$0.02\% \\
\midrule
\multicolumn{10}{c}{\textit{PEMS-BAY}} \\
\cmidrule{1-10}
PDFormer & 1.33$\pm$0.00 & 2.81$\pm$0.01 & 2.83$\pm$0.04\% & 1.65$\pm$0.01 & 3.75$\pm$0.02 & 3.73$\pm$0.06\% & 1.91$\pm$0.01 & 4.41$\pm$0.03 & 4.48$\pm$0.06\% \\
STAEformer & 1.33$\pm$0.01 & 2.81$\pm$0.02 & 2.83$\pm$0.02\% & 1.64$\pm$0.02 & \textbf{3.70$\pm$0.02} & 3.73$\pm$0.03\% & 1.91$\pm$0.02 & 4.35$\pm$0.03 & 4.51$\pm$0.04\% \\
TESTAM & \textbf{1.30$\pm$0.00} & \textbf{2.79$\pm$0.02} & \textbf{2.75$\pm$0.03\%} & \textbf{1.60$\pm$0.01} & \textbf{3.70$\pm$0.04} & \textbf{3.62$\pm$0.05\%} & \textbf{1.85$\pm$0.01} & \textbf{4.34$\pm$0.04} & \textbf{4.35$\pm$0.02\%} \\
\textbf{SAGT-F (ours)} & 1.37$\pm$0.01 & 2.90$\pm$0.05 & 2.96$\pm$0.05\% & 1.70$\pm$0.04 & 3.89$\pm$0.11 & 3.90$\pm$0.11\% & 1.99$\pm$0.05 & 4.60$\pm$0.12 & 4.67$\pm$0.12\% \\
GC-MoE$^{\dagger}$ & 1.32$\pm$0.00 & 2.76$\pm$0.00 & 2.79$\pm$0.00\% & 1.63$\pm$0.00 & 3.69$\pm$0.00 & 3.71$\pm$0.00\% & 1.92$\pm$0.00 & 4.38$\pm$0.00 & 4.55$\pm$0.00\% \\
\midrule
\multicolumn{10}{c}{\textit{PEMS07}} \\
\cmidrule{1-10}
PDFormer & 21.58$\pm$0.79 & 32.50$\pm$0.57 & 14.63$\pm$0.94\% & 22.74$\pm$0.53 & 34.97$\pm$0.34 & 14.12$\pm$0.31\% & 25.50$\pm$1.06 & 38.98$\pm$0.94 & 15.84$\pm$1.10\% \\
STAEformer & 19.16$\pm$0.06 & 31.30$\pm$0.07 & 8.05$\pm$0.04\% & 20.50$\pm$0.09 & 33.82$\pm$0.09 & 8.61$\pm$0.07\% & 22.73$\pm$0.16 & 37.53$\pm$0.12 & 9.75$\pm$0.15\% \\
TESTAM & \textbf{18.79$\pm$0.13} & \textbf{31.11$\pm$0.27} & \textbf{7.86$\pm$0.09\%} & \textbf{20.09$\pm$0.16} & \textbf{33.57$\pm$0.33} & \textbf{8.38$\pm$0.08\%} & \textbf{22.00$\pm$0.14} & \textbf{36.76$\pm$0.40} & \textbf{9.23$\pm$0.06\%} \\
\textbf{SAGT-F (ours)} & 19.16$\pm$0.04 & 31.34$\pm$0.09 & 8.05$\pm$0.05\% & 20.63$\pm$0.04 & 34.05$\pm$0.16 & 8.62$\pm$0.02\% & 22.82$\pm$0.06 & 37.79$\pm$0.26 & 9.60$\pm$0.07\% \\
GC-MoE$^{\dagger}$ & 18.89$\pm$0.01 & 31.10$\pm$0.01 & 8.08$\pm$0.06\% & 20.20$\pm$0.01 & 33.72$\pm$0.00 & 8.58$\pm$0.04\% & 22.09$\pm$0.02 & 36.98$\pm$0.01 & 9.53$\pm$0.01\% \\
\midrule
\multicolumn{10}{c}{\textit{EXPY-TKY}} \\
\cmidrule{1-10}
PDFormer & 6.73$\pm$0.01 & 10.58$\pm$0.03 & 30.57$\pm$0.52\% & 7.20$\pm$0.04 & 11.48$\pm$0.02 & 34.19$\pm$0.83\% & 7.66$\pm$0.04 & 12.25$\pm$0.04 & 36.83$\pm$1.08\% \\
STAEformer & 6.56$\pm$0.03 & 10.44$\pm$0.02 & 29.79$\pm$0.26\% & 6.86$\pm$0.03 & 11.10$\pm$0.06 & 31.78$\pm$0.27\% & 7.17$\pm$0.02 & 11.70$\pm$0.04 & 33.64$\pm$0.41\% \\
TESTAM & \textbf{6.49$\pm$0.04} & \textbf{10.43$\pm$0.08} & \textbf{29.39$\pm$0.14\%} & \textbf{6.78$\pm$0.06} & \textbf{11.03$\pm$0.10} & \textbf{31.45$\pm$0.23\%} & \textbf{7.09$\pm$0.09} & \textbf{11.68$\pm$0.16} & 32.87$\pm$0.39\% \\
\textbf{SAGT-F (ours)} & 6.66$\pm$0.01 & 10.57$\pm$0.04 & 29.48$\pm$0.14\% & 6.92$\pm$0.02 & 11.20$\pm$0.03 & 31.93$\pm$0.50\% & 7.18$\pm$0.05 & 11.70$\pm$0.10 & \textbf{32.78$\pm$0.30\%} \\
GC-MoE$^{\dagger}$ & 6.46$\pm$0.00 & 10.25$\pm$0.00 & 29.69$\pm$0.01\% & 6.78$\pm$0.01 & 10.88$\pm$0.01 & 31.48$\pm$0.02\% & 7.16$\pm$0.01 & 11.54$\pm$0.02 & 33.48$\pm$0.06\% \\
\bottomrule
\end{tabular}
\end{table*}

\subsection{Comparison With Composite Forecasters}
\label{sec:composite}

For completeness, we compare against GC-MoE~\cite{gcmoe}, which addresses the same task by a different route: instead of training a single model, it composes three frozen, pretrained forecasters via a learned router. We report it separately; note that this distinction is not simply that it employs a mixture of experts, since TESTAM~\cite{lee2024testam} does so as well, but that TESTAM trains its experts jointly from random initialization, whereas GC-MoE presupposes them.

As Table~\ref{tab:main} shows, GC-MoE achieves lower error than SAGT-F on five of six benchmarks, by margins of $0.06$ to $0.40$~MAE (largest on PEMS07), while SAGT-F is more accurate only on PEMS08, by $0.03$, which is within seed variation. GC-MoE is also by far the most seed-stable entry in the table (standard deviation of at most $0.01$~MAE), because only its router is retrained per seed, while the three experts remain fixed. This is an unsurprising outcome: an ensemble of three independently trained forecasters has more effective capacity than any single model, and composite methods are generally expected to outperform their components. The relevant comparison is therefore one of cost. Once expert pretraining is counted, GC-MoE requires more training computation than SAGT-F, and it cannot be deployed at all without first obtaining three separate forecasters. SAGT-F reaches comparable accuracy as a single end-to-end model with no pretrained components.

\subsection{Incremental Construction}
\label{sec:ladder}

SAGT is evaluated as a sequence of controlled modifications, each isolating one design decision (Table~\ref{tab:ladder}). Variants A--C retain the idea of temporal machinery (a DFT--convolution cell) and differ only in how the spatial operator is constructed; variants D--F adopt the transformer temporal backbone.

\begin{table}[!t]
\caption{Incremental Construction of SAGT on PEMS04. Each Row Adds One Modification to the Row Above, Except Row F, Which Introduces Both the Directed Two-Basis Operator and the Local Branch; These Are Separated in Table~\ref{tab:ablation}}
\label{tab:ladder}
\centering
\tiny
\begin{tabular}{clccc}
\toprule
Var. & Modification introduced & MAE & RMSE & MAPE \\
\midrule
A & Learned adjacency & 25.96$\pm$0.86 & 39.39$\pm$1.42 & 24.06$\pm$2.67\% \\
B & Learned orthonormal basis & 25.08$\pm$0.50 & 38.80$\pm$0.62 & 18.07$\pm$0.70\% \\
C & Input-conditioned filter & 25.05$\pm$0.56 & 38.63$\pm$0.63 & 18.45$\pm$0.78\% \\
D & Transformer temporal backbone & 19.14$\pm$0.19 & 31.12$\pm$0.30 & 13.38$\pm$0.22\% \\
E & \quad + frequency-bin attention & 19.04$\pm$0.13 & 30.98$\pm$0.31 & 13.10$\pm$0.01\% \\
F & \quad + directed 2-basis, local (full) & 18.89$\pm$0.13 & 30.52$\pm$0.33 & 13.14$\pm$0.20\% \\
\bottomrule
\end{tabular}

\end{table}

Every row that uses a spectral basis (B onward) is reported at the rank $k=32$ used by the full model, so that consecutive rows differ only in the stated modification. Two observations follow. First, replacing the Chebyshev approximation of the graph-Fourier basis with a directly learned orthonormal basis (A$\to$B) improves MAE by $0.88$ while removing the eigendecomposition from the pipeline. We do not read this as evidence that the analytic basis is inferior, since this row also introduces the rank-32 truncation and the two effects are not separated. However, we observe that a learned basis is an adequate replacement obtained at lower cost. Making the filter input-conditioned (B$\to$C) changes MAE by only $0.03$, far inside the seed variation of these three rows (standard deviation $0.5$--$0.9$). Second, by far the largest single improvement is the replacement of the DFT--convolution temporal cell by the transformer backbone (C$\to$D, $-5.91$~MAE), an order of magnitude more than any other step and consistent with the strength reported for adaptive embeddings~\cite{liu2023staeformer}.

The third observation concerns the two components proposed in this work. Adding frequency-bin attention to the single-basis operator improves performance on the PEMS04 dataset by $0.10$~MAE (D$\to$E); replacing that operator with the directed hybrid one, with frequency attention now present on both sides of the comparison, gives a further $0.15$ (E$\to$F), for $0.25$ in total. Each step on its own is comparable to the seed standard deviation of these rows ($0.13$--$0.19$).

\subsection{Component Ablations}
\label{sec:ablation}

Table~\ref{tab:ablation} removes one component at a time from the full model. On the PEMS08 dataset, the two spatial branches carry the gain ($+0.17$ and $+0.25$~MAE when the local and global branches are removed), whereas removing frequency attention leaves MAE unchanged ($-0.01$, well inside the $\pm0.29$ spread of that configuration). On PEMS04, the ordering reverses: frequency attention is the largest single contributor ($+0.36$), the global branch adds $+0.19$, and removing the local diffusion has no measurable effect. The global low-rank operator is thus the only component whose removal hurts both datasets; the importance of the other two is dataset-dependent.

\begin{table}[!t]
\renewcommand{\arraystretch}{1.15}
\caption{Component Ablations of the Full Model (Higher Is Worse).}
\label{tab:ablation}
\centering
\footnotesize
\begin{tabular}{lcccc}
\toprule
\multirow{2}{*}{Configuration} & \multicolumn{2}{c}{PEMS04} & \multicolumn{2}{c}{PEMS08} \\
\cmidrule(lr){2-3}\cmidrule(lr){4-5}
 & MAE & $\Delta$ & MAE & $\Delta$ \\
\midrule
SAGT-F (full) & 18.89$\pm$0.13 & -- & 14.39$\pm$0.11 & -- \\
w/o local diffusion & 18.89$\pm$0.05 & $0.00$ & 14.56$\pm$0.12 & $+0.17$ \\
w/o global spectral & 19.08$\pm$0.23 & $+0.19$ & 14.64$\pm$0.07 & $+0.25$ \\
w/o freq. attention & 19.25$\pm$0.35 & $+0.36$ & 14.38$\pm$0.29 & $-0.01$ \\
\bottomrule
\end{tabular}
\end{table}

\begin{table}[!t]
\renewcommand{\arraystretch}{1.15}
\caption{Rank Parameter (K) Sweep Experiments (Higher Is Worse).}
\label{tab:k-sweep}
\centering
\footnotesize
\begin{tabular}{lcccc}
\toprule
\multirow{2}{*}{Configuration} & \multicolumn{2}{c}{PEMS04} & \multicolumn{2}{c}{PEMS08} \\
\cmidrule(lr){2-3}\cmidrule(lr){4-5}
 & MAE & $\Delta$ & MAE & $\Delta$ \\
\midrule
k = 16 & 19.04$\pm$0.09 & $+0.15$ & 14.39$\pm$0.09 & $0.00$ \\
k = 32 (default) & 18.89$\pm$0.13 & -- & 14.39$\pm$0.11 & -- \\
k = 64 & 18.92$\pm$0.08 & $+0.03$ & 14.49$\pm$0.07 & $+0.10$ \\
\bottomrule
\end{tabular}
\end{table}

\subsubsection{Spectral Rank}
The rank $k$ sets how many spatial modes the operator retains and is the single capacity parameter of the spectral branch. Table~\ref{tab:k-sweep} varies it in the full model on the PEMS04 and PEMS08 datasets. Halving the rank to $k=16$ costs $0.15$~MAE on PEMS04 and leaves performance on PEMS08 unchanged; doubling it to $k=64$ is slightly worse on both ($+0.03$ and $+0.10$). No alternative improves on $k=32$ on either dataset, and cost remains linear in $N$ throughout.

We fix $k=32$ across all variants and datasets rather than selecting it per benchmark for three reasons. First, the useful rank plausibly depends on the size and structure of the sensor network, which varies by an order of magnitude across our benchmark datasets. Second, selecting $k$ per dataset would tune a capacity parameter for each benchmark while every baseline runs at its published default, which would flatter our results rather than isolate the architecture's contribution. Third, a rank that changed between rows of Table~\ref{tab:ladder} would make each row differ in two respects at once, so the modification named beside it could no longer be read as the cause of the change. A uniform $k=32$ keeps all three comparisons clean, and the sweep shows that it costs no accuracy on the two datasets where it was measured. Analysis of how the useful rank scales with network size beyond these two datasets is left to future work.

\subsection{Efficiency and Scalability}
\label{sec:efficiency}

We analyze computational costs on the EXPY-TKY dataset, the largest network, under two regimes. Table~\ref{tab:eff} fixes the batch size at $1$, where every model fits, giving a controlled comparison, and at $8$, which reveals the practical training limit. All measurements are taken on a single NVIDIA Tesla P100 ($16$~GB) after discarding warm-up iterations.

\begin{table*}[!t]
\renewcommand{\arraystretch}{1.2}
\caption{Computational Cost on EXPY-TKY ($N{=}1843$). All Columns, Including Parameter Counts, Are Measured on This Benchmark; Models With Node Embeddings Grow With $N$, So These Counts Are Not Comparable With Values Reported on Smaller Networks. Per-Iteration Figures Are Measured at Two Batch Sizes; OOM Denotes Exhaustion of $16$~GB.}
\label{tab:eff}
\centering
\begin{tabular}{lccccccc}
\toprule
\multirow{2}{*}{Model} & \multirow{2}{*}{Params}
 & \multicolumn{3}{c}{Batch size 1 (all models fit)} & \multicolumn{3}{c}{Batch size 8} \\
\cmidrule(lr){3-5}\cmidrule(lr){6-8}
 & & Mem.\ (GB) & Train (ms) & Infer (ms) & Mem.\ (GB) & Train (ms) & Infer (ms)\\
\midrule
STGCN      & 1.08\,M  & 0.15 & 10.9  & 4.0   & 0.61  & 47.5   & 16.5 \\
DCRNN      & 0.37\,M  & 0.54 & 372.8 & 104.4 & 4.17  & 647.7  & 269.4 \\
StemGNN    & 11.33\,M & 0.53 & 636.9 & 214.6 & 2.53  & 670.8  & 284.0 \\
MegaCRN    & 0.45\,M  & 4.29 & 987.7 & 466.5 & 7.50  & 1650.5 & 672.6 \\
AGCRN      & 0.76\,M  & 9.47 & 419.9 & 239.7 & 10.65 & 586.4  & 247.7 \\
STAEformer & 2.83\,M  & 4.79 & 212.6 & 73.7  & OOM & -- & -- \\
TESTAM     & 0.32\,M  & 5.04 & 241.6 & 86.5  & OOM & -- & -- \\
\midrule
SAGT-D     & 3.15\,M  & 0.94 & 72.4  & 19.4  & 7.14 & 500.9 & 136.8 \\
\textbf{SAGT-F (ours)} & 3.25\,M & 1.07 & 97.5 & 29.0 & 8.22 & 665.5 & 205.4 \\
\midrule
\multicolumn{8}{l}{\footnotesize\emph{Composite method (different training protocol)}}\\
GC-MoE & 2.25\,M & 0.69 & 235.9 & 224.4 & 1.00 & 391.9 & 383.0 \\
\bottomrule
\end{tabular}
\end{table*}

At an identical batch size, SAGT-F requires $4.5\times$ less memory than STAEformer and $4.7\times$ less than TESTAM, while training $2.2\times$ and $2.5\times$ faster, respectively, and performing inference $2.5\times$ and $3.0\times$ faster. The decisive detail is that SAGT-F has \emph{more} parameters than STAEformer ($3.25$\,M versus $2.83$\,M), yet occupies far less memory. Two caveats attach to the timing columns. Batch size $1$ under-utilizes the accelerator, so per-iteration times in that regime are influenced by kernel-launch overhead and should be read as a controlled comparison rather than as achievable throughput. The quantity a practitioner experiences is throughput at each model's largest feasible batch size, which on this benchmark is $8$ for SAGT-F and $1$ for STAEformer and TESTAM: $83.2$ms per sample against $212.6$ and $241.6$ respectively, a $2.6 \times$ -- $2.9 \times$ difference that follows directly from the memory profile rather than from arithmetic cost. The saving, therefore, cannot be attributed to a smaller model; it follows from the spatial operator never materializing an $N \times N$ activation, exactly as predicted by the analysis in Section~\ref{sec:complexity}.

This headroom translates into a practical difference, and it is not an artifact of the benchmark harness. At batch size $8$, both leading attention-based baselines exhaust memory, whereas SAGT trains comfortably.

\subsection{Inspecting the Learned Frequency Weights}
\label{sec:freqanalysis}

\begin{figure*}[!t]
\centering
\includegraphics[width=0.85\textwidth]{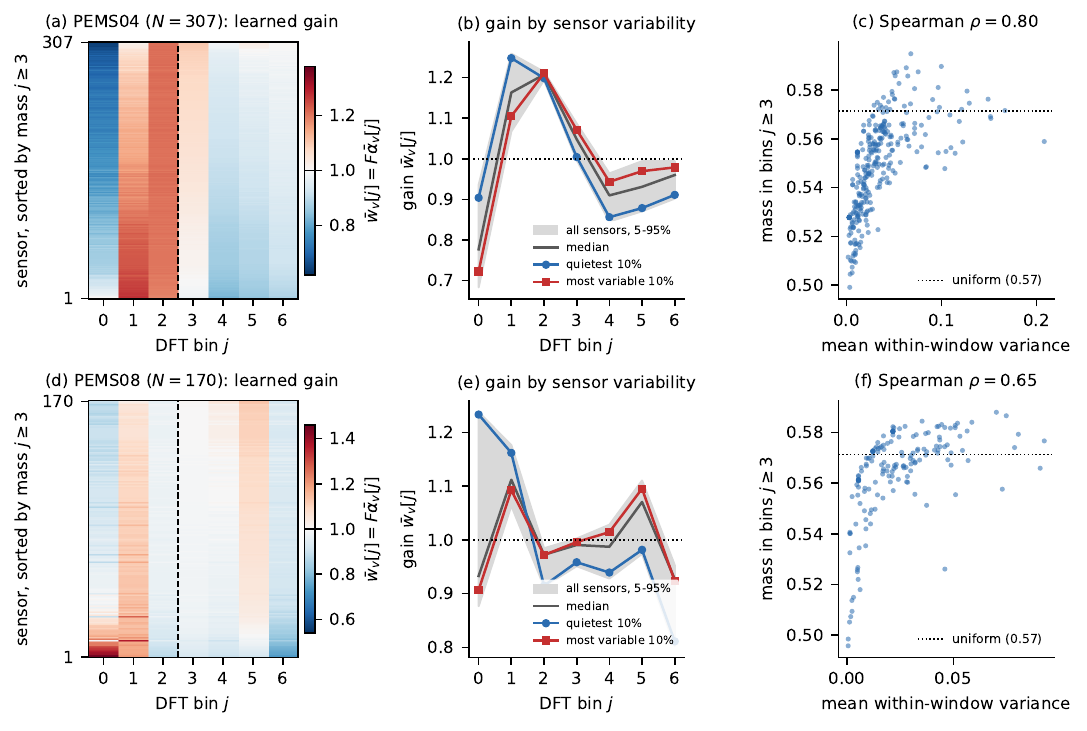}
\caption{Frequency-bin attention averaged over the test split, in unit-mean gain units $\bar{w}_v[j]=F\bar{\alpha}_v[j]$ ($1.0$ is the identity). (a,d) Per-sensor gain, sensors sorted by mass in bins $j\geq3$. (b,e) Gain per bin for the quietest and most variable sensor deciles, with the $5$--$95$th percentile band. (c,f) Mass in bins $j\geq3$ against mean within-window variance, with Spearman $\rho$; dotted line marks the uniform value $4/7$. Top: PEMS04 ($N=307$). Bottom: PEMS08 ($N=170$).}
\label{fig:freqattn}
\end{figure*}

Figure~\ref{fig:freqattn} shows the learned weights averaged over the test split, in unit-mean gain units where $1.0$ is the identity. The gains are modest; the decile means span $0.72$ to $1.25$, so the module reweights rather than selects periodicities. The profile tilts with sensor variability: per-bin Spearman correlations against mean within-window variance run from $-0.81$ at $j=1$ to $+0.81$ at $j=5$, and aggregated over $j\geq3$, the relationship is monotone but saturating ($\rho=0.80$ on PEMS04, $0.65$ on PEMS08). High-frequency content is thus suppressed most on the quietest sensors and less as a sensor becomes more variable; since the weights lie on the simplex, the opposing signs describe one tilt rather than two. The ablation removes this reweighting together with the conditioning path that produces it, so we report the figure as a consistency check on the learned weights, from a single checkpoint, rather than as evidence of a physical mechanism.

\subsection{Discussion}
\label{sec:discussion}

Taken together, the results demonstrate that SAGT shows competitive accuracy: it achieves the best result on one of six benchmarks (PEMS08), is within $0.13$~MAE of the leader on METR-LA, PEMS-BAY and EXPY-TKY, and ranks third on PEMS04 and PEMS07, trailing the leader by $0.66$ and $0.52$~MAE, respectively. However, what distinguishes it is that this accuracy is achieved at roughly a quarter of the memory and half the training time of the models it matches, and at a scale where those models are no longer trainable on commodity hardware at a practical batch size. That comparison is against the dense-attention family, which is where the accuracy frontier on these benchmarks lies.

The comparison with a composite forecaster in Section~\ref{sec:composite} supports the same cost-centered conclusion. An ensemble of three pretrained experts achieves lower error on five of six benchmarks, as expected from its greater effective capacity, but only after incurring the cost of three separate training runs. Once that cost is accounted for, SAGT-F is cheaper to obtain on the largest benchmark, remains within $0.40$~MAE of it on any benchmark and within $0.11$ on the largest, and is level with it on PEMS08.

The ablations clarify how the two mechanisms introduced here should be understood. On PEMS04, removing frequency attention from the full model costs $0.36$~MAE, whereas removing local diffusion costs nothing. On PEMS08, removing frequency attention costs almost nothing, whereas removing local diffusion costs $0.17$~MAE. Removing the global spectral branch hurts both datasets. The global low-rank operator is therefore the core of the spatial layer, while local diffusion and frequency-bin attention act as complementary corrections that address different sources of error, local propagation that a rank-k operator cannot express and sensor-level within-window noise, and which of them matters depends on the network. This dataset dependence is also why a fixed combination of the two may leave accuracy unused. 

\section{Conclusion}
\label{sec:conclusion}

We introduced the Spectral Adaptive Graph Transformer (SAGT), a traffic forecasting model whose spatial operator is learned, directed, and input-conditioned, and whose cost grows linearly rather than quadratically with the number of sensors. The operator $\mathbf{U}_q\operatorname{diag}(\boldsymbol{\theta}_{b,t})\mathbf{U}_k^{\!\top}$ replaces the Laplacian eigenbasis of classical spectral graph networks~\cite{cao2020spectral,defferrard2016convolutional} with a pair of learned orthonormal bases of rank $k\ll N$, eliminating the $\mathcal{O}(N^3)$ eigendecomposition, permitting asymmetric influence between upstream and downstream sensors, and allowing the filter to change with traffic conditions. A sparse road-graph diffusion branch supplies the strictly local propagation that a low-rank operator cannot express, and a zero-initialized gate learns the balance between the two. Ahead of spatial mixing, unit-mean gain frequency-bin attention reweights each sensor's discrete Fourier bins according to its identity and current context.

Across six benchmarks spanning $170$ to $1843$ sensors, SAGT achieves the best accuracy of any single end-to-end model on PEMS08, and is within $0.13$~MAE of the leader on METR-LA, PEMS-BAY and EXPY-TKY, and is third on PEMS04 and PEMS07, trailing the leader by $0.66$ and $0.52$~MAE, respectively; outside PEMS08, it is competitive but not leading. Its distinguishing property is therefore not accuracy but the cost at which comparable accuracy is obtained: at an identical batch size on a $1843$-sensor network, it uses approximately one quarter of the memory of STAEformer~\cite{liu2023staeformer} and TESTAM~\cite{lee2024testam} while training roughly twice as fast, and it remains trainable at batch sizes where both exhaust a $16$\,GB GPU. Since SAGT has more parameters than STAEformer yet occupies substantially less memory, the saving is attributable to the architecture, since no $N\times N$ activation is ever materialized, rather than to reduced capacity.

Several limitations bound these conclusions. SAGT's ablations show that the global and local branches contribute unequally across datasets, suggesting that the fixed architecture may be leaving accuracy on the table in networks whose structure favors one branch. The SAGT's architecture has only been tested with traffic datasets, and its effectiveness has not been evaluated on more diverse spatio-temporal prediction tasks. Component ablations and parameter analyses are limited in this study; more comprehensive experiments are needed to better understand their effects and their relationships with each dataset's characteristics.

Future work follows naturally from these observations. The spectral rank $k$ is held fixed at a single value across all six benchmarks, deliberately, so that no comparison is confounded by capacity; the sweep in Section~\ref{sec:ablation} shows that neither halving nor doubling it helps the full model on PEMS04 or PEMS08, but how the useful rank scales with network size is left open. Adapting $k$ rather than fixing it is a natural extension. The interpretability afforded by the frequency-bin attention weights and the learned spatial modes has not been exploited analytically here, and relating learned modes to physical corridors would be a substantive study in itself. More broadly, the spatial operator makes no assumption specific to road traffic, and evaluating it on other large-scale spatio-temporal domains, power grids, air quality networks, or sensor networks generally would test whether the efficiency advantage reported here generalizes.

\section*{Data and Code Availability}
All six datasets are publicly available from their original sources. The implementation is available at \url{https://github.com/Ahghaffari/sagt}.

\ifCLASSOPTIONcaptionsoff
  \newpage
\fi

\bibliographystyle{IEEEtran}
\bibliography{ref}

\end{document}